\documentclass[lettersize,journal]{IEEEtran}
\usepackage{amsmath,amsfonts,multirow}
\usepackage{algorithm}
\usepackage{algorithmicx}
\usepackage{algpseudocode}
\usepackage{array}
\usepackage{textcomp}
\usepackage{stfloats}
\usepackage{url}
\usepackage{verbatim}
\usepackage{graphicx}
\usepackage{cite}
\usepackage{bm}
\usepackage{amssymb}
\usepackage{bbding} 
\usepackage{booktabs}
\usepackage{makecell}
\usepackage{color}
\usepackage[table]{xcolor}
\usepackage{subcaption}

\usepackage[implicit=false]{hyperref}
\usepackage{comment}

\def\etal{\emph{et al.}}

\newlength\savewidth
  
\begin{document}

\title{Semi-LLIE: Semi-supervised Contrastive Learning with Mamba-based Low-light Image Enhancement}

\author{Guanlin~Li, Ke~Zhang, Ting~Wang, Ming~Li, Bin~Zhao, and Xuelong~Li,~\IEEEmembership{Fellow,~IEEE}
	
	\IEEEcompsocitemizethanks{
		
		\IEEEcompsocthanksitem 
        Guanlin Li is with the School of Computer Science and also with the School of Artificial Intelligence, Optics, and Electronics (iOPEN), Northwestern Polytechnical University, Xi'an 710072, P.R. China. 

        Ke Zhang and Ting Wang are with Science and Technology on Electromechanical Dynamic Control Laboratory, Xi’an, P.R. China.

        Ming Li is with the School of Astronautics, Northwestern Polytechnical University, Xi'an P.R. China, and also with  Science and Technology on Electromechanical Dynamic Control Laboratory, Xi’an, China.
        
		Bin Zhao and Xuelong Li are with the School of Artificial Intelligence, OPtics and ElectroNics (iOPEN), Northwestern Polytechnical University, Xi'an 710072, P.R. China, and also with the Key Laboratory of Intelligent Interaction and Applications (Northwestern Polytechnical University), Ministry of Industry and Information Technology, Xi'an 710072, P. R. China. This work was partly supported by the National Natural Science Foundation of China under Grant 62376222. (\textit{Corresponding author: Ke Zhang}) (E-mail: liguanlin1229@gmail.com; 32578792@qq.com).
		
	}	

}
\maketitle

\begin{abstract}
Despite the impressive advancements made in recent low-light image enhancement techniques, the scarcity of paired data has emerged as a significant obstacle to further advancements. This work proposes a mean-teacher-based semi-supervised low-light enhancement (Semi-LLIE) framework that integrates the unpaired data into model training. The mean-teacher technique is a prominent semi-supervised learning method, successfully adopted for addressing high-level and low-level vision tasks. However, two primary issues hinder the naive mean-teacher method from attaining optimal performance in low-light image enhancement. Firstly,  pixel-wise consistency loss is insufficient for transferring realistic illumination distribution from the teacher to the student model, which results in color cast in the enhanced images. Secondly, cutting-edge image enhancement approaches fail to effectively cooperate with the mean-teacher framework to restore detailed information in dark areas due to their tendency to overlook modeling structured information within local regions. To mitigate the above issues, we first introduce a semantic-aware contrastive loss to faithfully transfer the illumination distribution, contributing to enhancing images with natural colors. 
Then, we design a Mamba-based low-light image enhancement backbone to effectively enhance Mamba's local region pixel relationship representation ability with a multi-scale feature learning scheme, facilitating the generation of images with rich textural details. 
Further, we propose novel perceptive loss based on the large-scale vision-language Recognize Anything Model (RAM) to help generate enhanced images with richer textual details.
The experimental results indicate that our Semi-LLIE surpasses existing methods in both quantitative and qualitative metrics.
The code and models are available at~\href{https://github.com/guanguanboy/Semi-LLIE}{https://github.com/guanguanboy/Semi-LLIE}.

\end{abstract}


\section{Introduction}

\IEEEPARstart{L}ow-light imaging refers to capturing images with insufficient photons arriving at the sensor, typically due to weak ambient light or short exposure time. It is inevitable in daily-life applications, such as nighttime photography, surveillance, and autonomous driving.
Low-light imaging significantly degrades image quality and severely hampers the performance improvement of downstream high-level tasks like object detection and segmentation. 

\begin{figure}[t]
  \centering
  \begin{subfigure}{0.49\linewidth}
    \includegraphics[width=1.\linewidth]{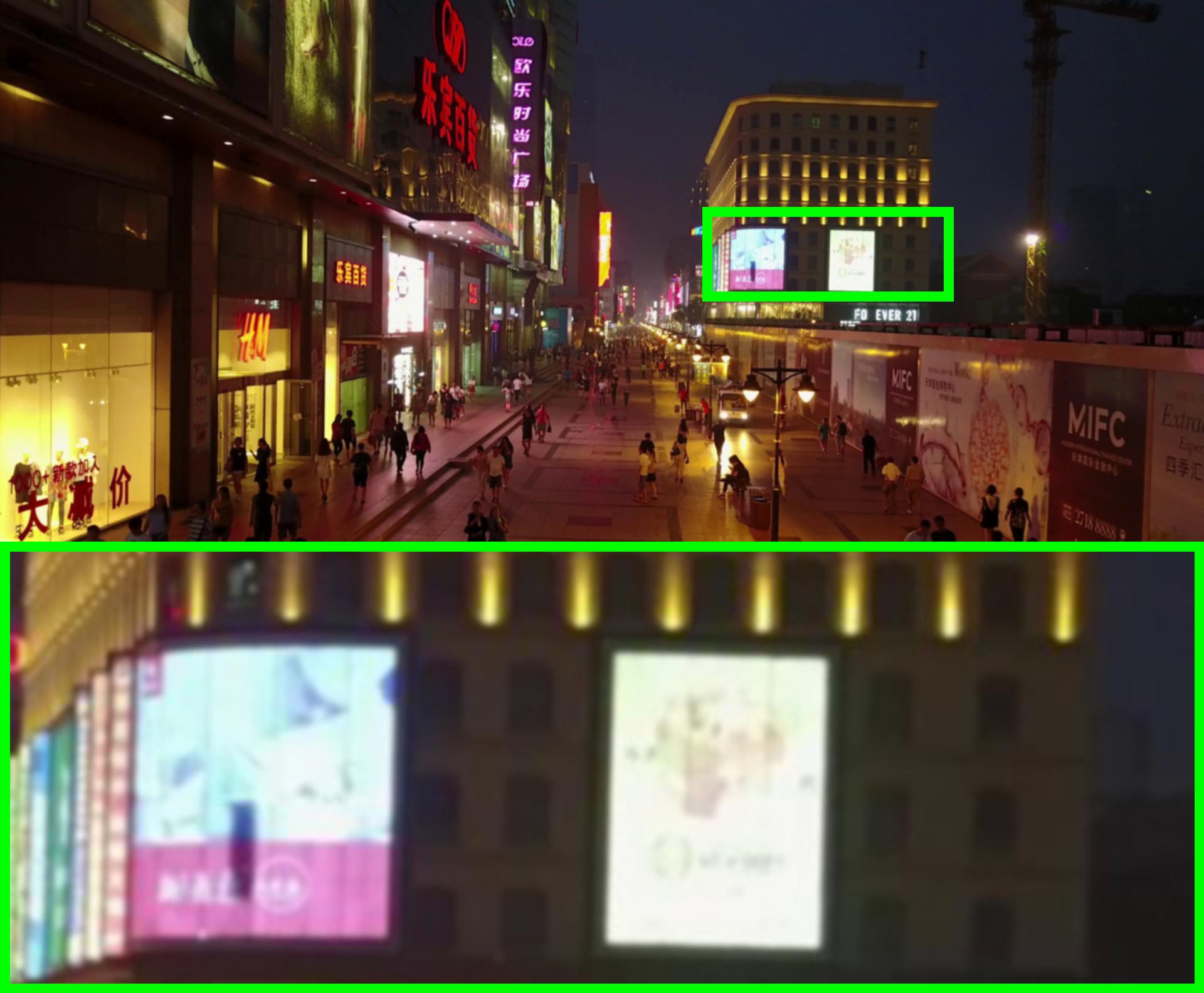}\vspace{-0.4em}

    \caption*{(a) Input}\medskip
    
    \includegraphics[width=1.\linewidth]{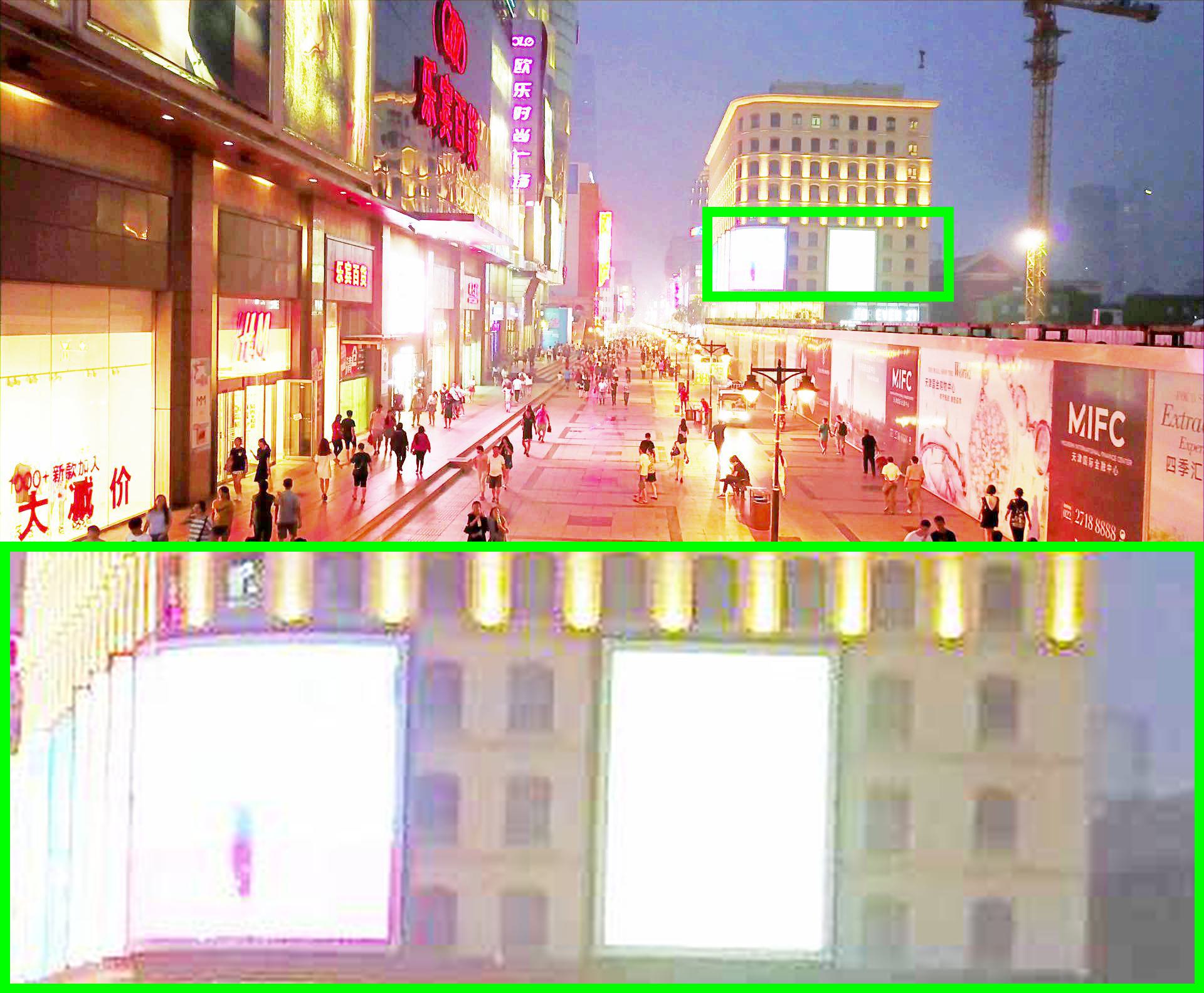}\vspace{-0.4em}
    
    \caption*{(c) SCI~\cite{ma2022sci}}
    
  \end{subfigure}
  \hfill
  \begin{subfigure}{0.49\linewidth}
    \includegraphics[width=1.\linewidth]{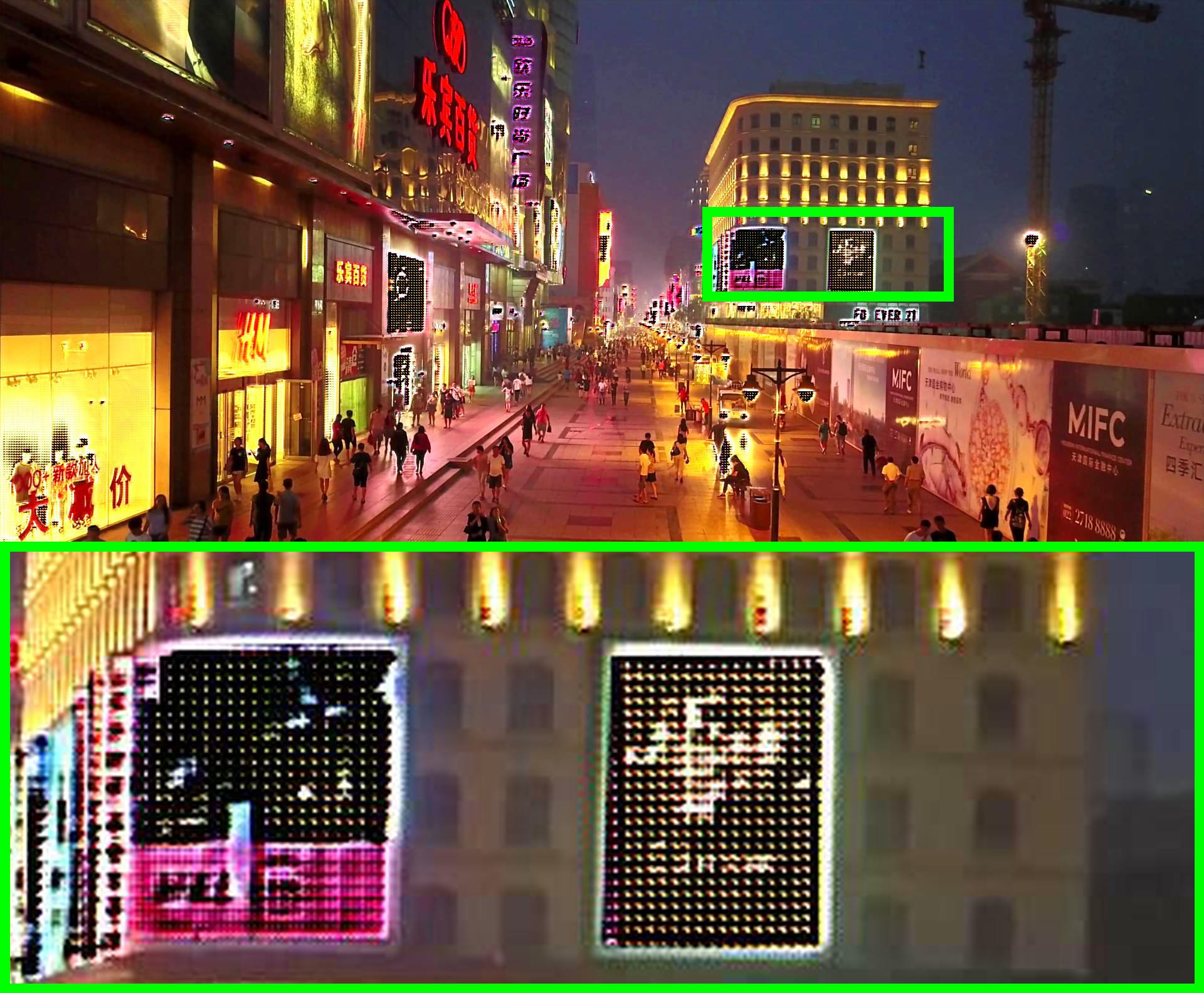}\vspace{-0.4em}

    \caption*{(b) Retinexformer~\cite{cai2023retinexformer}}\medskip
    
    \includegraphics[width=1.\linewidth]{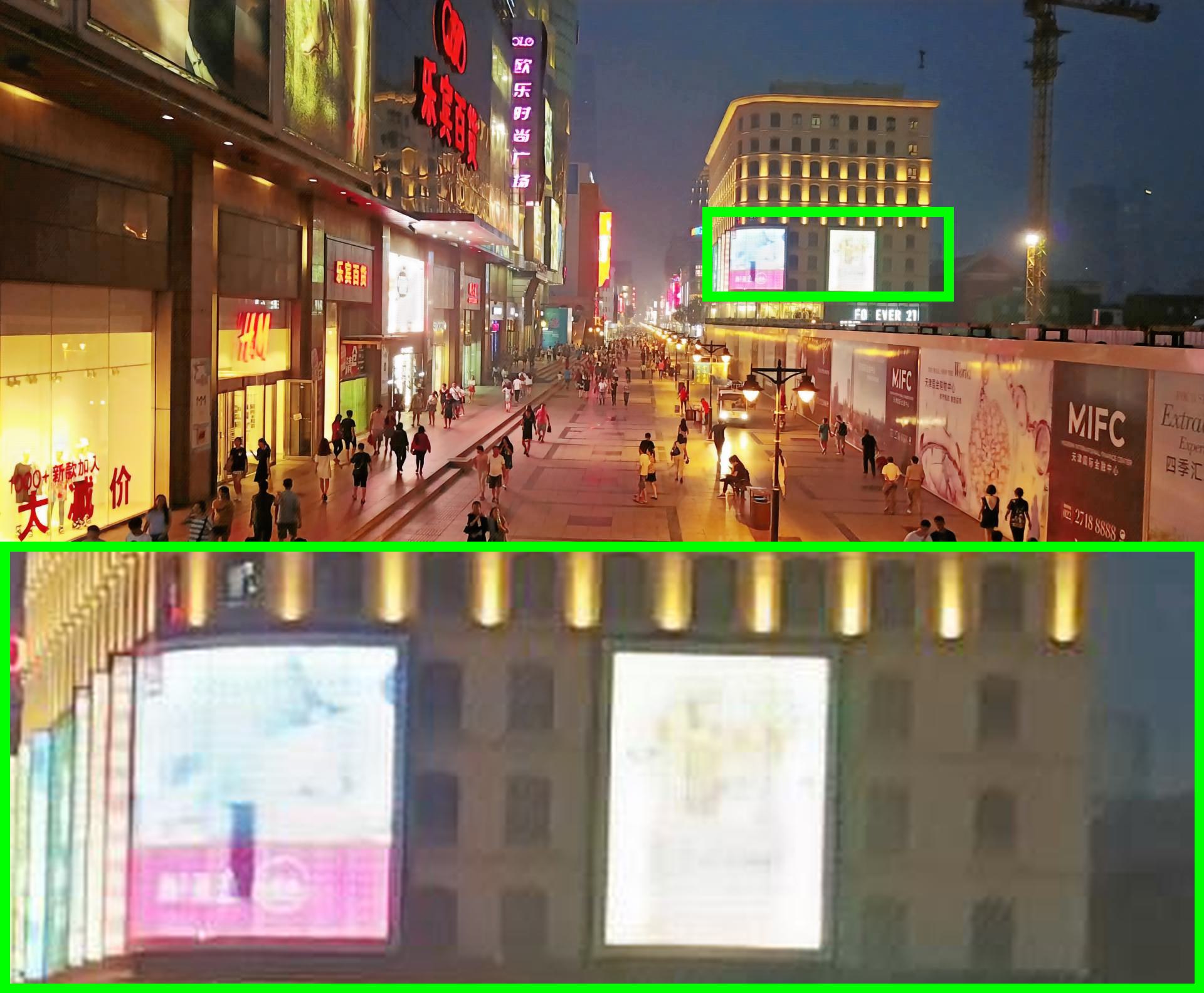}\vspace{-0.4em}
    
    \caption*{(d) Semi-LLIE (Ours)}
    
  \end{subfigure}

  \caption{Comparison with state-of-the-art (SOTA) supervised method Retinexformer~\cite{cai2023retinexformer} and unsupervised method SCI~\cite{ma2022sci} on the Visdrone dataset. It can be observed that the image enhanced by our method exhibits high contrast and visual-friendly textures. Other methods suffer from either textural detail distortion (b) or color deviation (c) as shown in the area marked by the \textbf{\textcolor{green}{green}} rectangles.}
  
  \label{fig:motivation}
  \vspace{-1.5em}
\end{figure}

The computational photography community has introduced numerous low-light image enhancement (LLIE) techniques~\cite{lin2023smnet, liu2021ruas, 10130403, lv2021attention, 10557144, jiang2021enlightengan} aimed at restoring images to normal illuminated levels.
Despite their achievements, most methods~\cite{9844872, 9894098} rely on a supervised learning paradigm which needs a substantial amount of paired data. However, obtaining paired low-light and normal images in real-world scenes is exceptionally challenging. 
One optional way to create paired datasets is synthesizing low-light images by changing the illumination of a normal-light image~\cite{Chen2018Retinex}. Another approach involves transforming norm-light images into low-light images with generative adversarial networks~\cite{goodfellow2014generative}. However, a substantial disparity exists between synthesized low-light images and images captured in real low-light scenarios. Consequently, models trained on synthetic datasets often struggle to generalize well in real-world scenes due to domain shifts. Besides, the existing paired benchmark datasets usually have a limited size. For instance, in the LOL-v1~\cite{Chen2018Retinex} and LOL-v2~\cite{Yang_Wang_Huang_Wang_Liu_2021} datasets, only $500$ and $789$ paired images are available. Models trained on such a small dataset are susceptible to overfitting. 

Different from creating paired image datasets, collecting unpaired low-light images is relatively straightforward. The main challenge lies in fully exploiting unpaired data's value.
To tackle this challenge, semi-supervised learning is one highly suitable technique, which leverages paired and unpaired data simultaneously for model optimization. 
This inspires us to develop a semi-supervised approach to boost the model's generalization capability for low-light image enhancement under real-world scenes.
Specifically, we employ the mean teacher approach~\cite{tarvainen2017mean} as the underlying framework. We choose the mean teacher method because it's one of the most effective semi-supervised techniques and has successfully addressed low-level vision challenges~\cite{liu2021synthetic, wang2022semi}.
It facilitates the acquisition of pseudo-labels for unpaired data and adopts a consistency loss to improve the network's robustness.

However, customizing the mean teacher method for low-light image enhancement poses significant challenges and requires careful consideration. Firstly, the commonly employed consistency loss in the mean teacher method is $L_1$ or $L_2$ reconstruction loss. This potentially leads to overfitting on incorrect predictions, which further introduces color distortion or textural artifacts in a low-light image. Secondly, precise local pixel dependencies and coherent global pixel relationships fulfill different and crucial functions for image enhancement~\cite{xu2022snr}. Local pixel dependencies are pivotal for the refinement of textures, while global pixel relationships are indispensable for accurately assessing the overall brightness level. Most SOTA low-light enhancement methods focus on modeling global brightness dependencies with Transformer-style self-attention~\cite{cai2023retinexformer, wang2022uformer} while neglecting to model local texture or illumination dependencies, making them less compatible with our Semi-LLIE framework. 

To mitigate the first issue, we incorporate semantic-aware contrastive loss as a regularization within our semi-supervised framework. 
Most contrastive regularizations are implemented within the latent feature space of a pre-trained object classification model~\cite{simonyan2015very}. They fail to incorporate fine-grained semantic structure understanding into their measurements. 
The recent emergence of the powerful image tagging model Recognize Anything Model (RAM)~\cite{zhang2024recognize} offers a novel approach to tackling these issues. RAM possesses powerful tagging capabilities and a strong zero-shot generalization capability that significantly outperforms CLIP~\cite{radford2021learning} and BLIP~\cite{li2022blip}.
It can output highly generic semantic information due to being trained on large-scale refined image-text pairs. 
We leverage the intermediate representation generated by RAM's image encoder to assess the semantic similarities between the original low-light images and their enhanced counterpart. Benefiting from the effective text-driven vision representations extracted from the image encoder of RAM, we can obtain an enhanced image with high contrast and natural colors (see Fig.~\ref{fig:motivation} (d)).

To alleviate the second issue, we propose a Mamba-based low-light image enhancement backbone to cooperate with our Semi-LLIE framework. Our enhancement backbone is capable of learning robust global-local pixel representations by integrating multi-scale local features with global features. It mainly consists of two stages.
%
The first stage is an illumination map estimation module consisting of several convolutional layers. The second stage is a Mamba-based image enhancement module integrating the Mamba and the multi-scale feature extractor.
Different from existing Mamba-based image restoration approaches~\cite{guo2024mambair}, where they neglect to focus on extracting multi-scale local textual features, the proposed Mamba-based low-light image enhancement backbone enhances the local feature representation of Mamba by designing multi-scale state space block (MSSB).
Besides, we develop a RAM-based perceptual loss leveraging the 
effective text-driven vision feature priors generated by different stages of the image encoder of RAM. 
It is utilized to optimize the enhancement model and improve the textual details of the enhanced images.

Extensive experimental results on the Visdrone~\cite{zhu2021detection} and LRSW~\cite{hai2023r2rnet} datasets show that the Semi-LLIE outperforms the current SOTA unsupervised methods by a large margin and even outperforms several influential supervised methods. Moreover, it enables us to generate enhanced images with rich local details and natural colors, further benefiting downstream object detection tasks. 
The main contributions of this work can be summarized as follows:
\begin{enumerate}
    \item We propose Semi-LLIE, a mean teacher-based semi-supervised low-light image enhancement framework, which effectively exploits the unpaired data to enhance generalization on real-world scenes.
    \item We design a RAM-based semantic-aware contrastive consistency loss benefiting from strong text-driven vision representations to help reduce color cast and generate visual-friendly enhanced images with natural colors.
    \item We design a Mamba-based low-light image enhancement backbone to integrate seamlessly with our Semi-LLIE. It learns robust local-global feature representations and enables producing images with rich details. Additionally, a RAM-based perceptual loss is proposed to further enhance the recovery of textural details.     
\end{enumerate}

\section{Related Works}

\subsection{Low-light Image Enhancement}

Deep learning-based low-light image enhancement techniques primarily aim to enhance the visual quality of images taken in low-light environments, ensuring that the enhanced images align with human visual perception~\cite{land1986alternative, jobson1997multiscale}. The majority of LLIE approaches~\cite{lv2021attention, liu2021ruas, wu2022uretinexnet, zhang2019kind} utilize a supervised learning framework, necessitating the availability of paired datasets for the training process. GAN-based methods~\cite{jiang2021enlightengan,10557144} eliminate the requirement for paired datasets in the training phase. However, their success heavily depends on the careful selection of unpaired data. Recently, curve-based methods~\cite{guo2020zerodce, STAR_ICCV} have been developed to enhance low-light images without relying on either paired or unpaired data. They achieve this by employing a set of carefully designed non-referenced loss functions. 
%
Recently, Kong~\etal~\cite{kong2024towards} have presented sequential learning and prompt learning strategies to tackle the multiple-in-one image enhancement problem.

\subsection{Semi-supervised Learning}
Semi-supervised learning approaches have gained prominence in computer vision, leveraging paired and unpaired data. Notable techniques include mean teacher~\cite{tarvainen2017mean}, virtual adversarial learning~\cite{miyato2018virtual}, and FixMatch~\cite{sohn2020fixmatch}, showcasing advancements in leveraging unpaired data for improved model performance.
Among these techniques, the mean teacher method~\cite{tarvainen2017mean} has demonstrated significant success in image recognition tasks.
The success of the mean teacher method in semi-supervised image recognition has spurred its application to other vision tasks, including semantic segmentation~\cite{hu2021semi, wang2022semi_semantic} and image restoration~\cite{liu2021synthetic, wang2022semi}.
As far as we know, the application of semi-supervised learning for enhancing low-light images has not been extensively studied.
Yang~\etal~\cite{yang2020drbn} takes an initial step by training a model using an integration of supervised and unsupervised losses for low-light image enhancement.
In contrast, our Semi-LLIE takes a more systematic approach by integrating mean teacher and contrastive loss to handle unpaired data in low-light image enhancement effectively.

\begin{figure*}[htbp]
	\centering
	\includegraphics[width=0.96\textwidth]{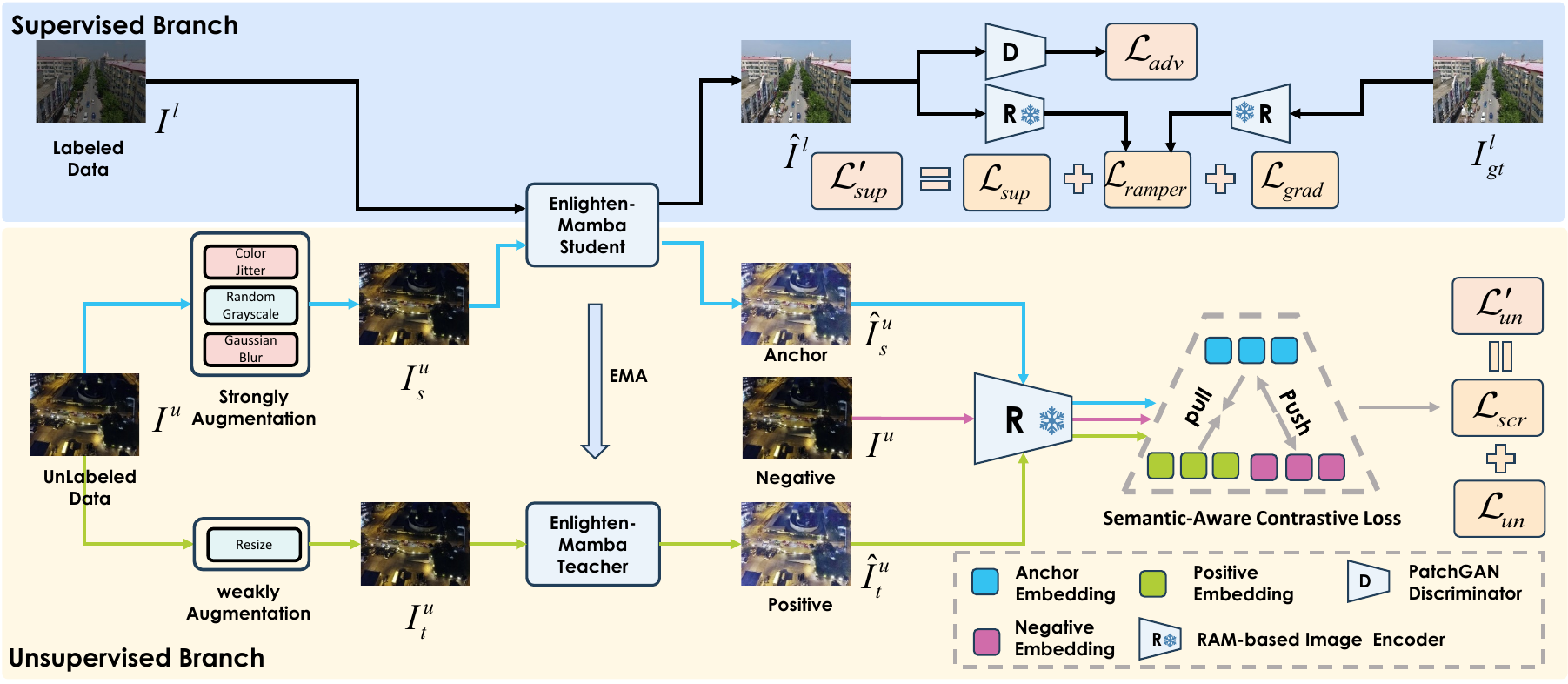}
	\caption[]{Architecture of our Semi-LLIE for low-light image enhancement. The Semi-LLIE employs the mean teacher paradigm, composed of a teacher and a student model. To faithfully transfer the illumination distribution from the teacher to the student model and reduce the color cast problem, we design a RAM-based semantic-aware contrastive loss $\mathcal{L}_{scr}$. To facilitate generating visual-friendly enhanced images with rich textural details, we design a RAM-based perceptual loss $\mathcal{L}_{ramper}$. The teacher model's weights are updated using the EMA from the student model. }
	\vspace{-0.2in}
	\label{fig:overall_framework}
\end{figure*}

\subsection{Contrastive Learning for Image Restoration}
Contrastive learning is a powerful paradigm in the field of self-supervised learning~\cite{chen2020simple, grill2020bootstrap, he2020momentum}. It facilitates the empirical learning of discriminative visual representations, ensuring that resembling samples are embedded closely whereas dissimilar samples are positioned farther apart. 
To harness the benefits of contrastive learning, previous studies in image restoration have primarily concentrated on constructing contrastive samples and refining the feature space.
For instance, in~\cite{liang2022semantically, wu2021contrastive}, clean and deteriorated images are treated as positive and negative. These images are then projected into a pre-defined feature space using VGG~\cite{simonyan2015very}.
It is worth noting that in the aforementioned studies, contrastive loss is employed within a supervised framework, which does not apply to unpaired data.
Making contrastive loss applicable to unpaired data remains an under-exploited challenge that requires further research and innovation.
In~\cite{liang2022semantically}, contrastive learning is introduced in the field of low-light image enhancement. However, it is important to acknowledge that this method is fundamentally a supervised learning approach, where contrastive loss serves as a regularization term, aiming to improve the efficacy of the supervised learning process.
%
In contrast, our work proposes a systematic semi-supervised approach to leverage contrastive learning for effectively utilizing unpaired data in low-light image enhancement.

\subsection{Vision State Space Models}

State Space Models (SSMs)~\cite{gu2021efficiently, gu2023mamba} has attracted great interest from searchers due to its long-range dependency modeling capability in linear complexity. Recently, Mamba has achieved promising performance in natural language processing with selective mechanisms and efficient hardware design. 
After that, vision Mambas are developed and adapted to a large range of tasks, including image classification~\cite{liu2024vmamba, zhu2024vision}, image restoration~\cite{guo2024mambair,bai2024retinexmamba}, and video analysis~\cite{wang2023selective}. 
Distinct from these outstanding works, we explore the potential of vision Mamba for low-light enhancement, incorporating multi-scale designs to enhance vision Mamba's local pixel relationship modeling capability.

\section{Methods}

\subsection{Preliminary}
Semi-supervised learning enables us to optimize a model using both paired and unpaired data.
Given a paired low-light dataset $D_l=\{(I^l,I_{gt}^l)\}$, $I^l \in \mathbb{R}^{H\times W\times 3}$ and $I_{gt}^l \in \mathbb{R}^{H\times W\times 3}$ represent the images sampled from the low-light set $\mathcal{I}_s^{Low}$ and normal-light set $\mathcal{I}_s^{Norm}$, respectively. Similarly, $I^u \in \mathbb{R}^{H\times W\times 3}$ represents a low-light image sampled from an unpaired data set $D_u=\{(I^u)\}, I^u\in \mathcal{I}_u^{Low}\}$.
Images in $D_l$ and $D_u$ do not overlap or have any duplication. We aim to design an image enhancement model to light up a low-light image and train it on both $D_l$ and $D_u$ simultaneously.

\subsection{Overall Structure} \label{sec:overall_structure}
The architecture of the proposed Semi-LLIE is depicted in Figure~\ref{fig:overall_framework}. It employs the classic structure of a mean-teacher-based semi-supervised learning framework~\cite{tarvainen2017mean, wang2022semi}. Our Semi-LLIE comprises two Mamba-based enhancement models with identical structures, the teacher and student models. The key difference between the two models is their strategy for updating weights. 

The weights $\omega_s$ of the student model are refined through direct application of the gradient descent algorithm.
The student model is usually optimized by minimizing the following objectives:
\begin{equation} \label{eq:loss_sec3.2}
\centering
\begin{aligned}
    \mathcal{L}_{s} = \mathcal{L}_{sup} + \alpha \mathcal{L}_{un},
\end{aligned}
\end{equation}
\noindent
where $\mathcal{L}_{sup} = ||f_s(I^l)-I_{gt}^l||_1$ refers to the objective for supervised optimization and $\mathcal{L}_{un} = ||f_s(I_{s}^u) - f_t(I_{t}^u)||_1$ denotes the consistency loss used to constrain the similarity between student and teacher model. $f_s$ and $f_t$ refer to the student and teacher enhancement model, respectively. $||\cdot||_1$ represent to $L_1$ loss. $I_{s}^u = \psi_s(I^u)$ and $I_{t}^u = \psi_t(I^u)$ are the augmented images to feed the student and teacher model. $\psi_s$ and $\psi_t$ represent the weak and strong data augmentation strategies of the students' and teachers' unpaired low-light input images. Weak augmentation refers to only applying resizing to the teacher's input image. While strong augmentation is implemented by imposing resizing, Gaussian blur, grayscale conversion, and color jittering, on the student's input image.

The weights $\omega_t$ of the teacher's model are updated by the calculation of the exponential moving average (EMA) of the student's weights $\omega_s$:
\begin{equation}\label{eq1}
    \omega_t=\beta \omega_t + (1-\beta) \omega_s,
\end{equation}
where $\beta\in(0,1)$ represent the moving average momentum. This strategy allows the teacher model to aggregate the updated weights directly following each optimization step. Besides, this is highly beneficial for stabilizing the training procedure and improving the model's performance~\cite{polyak1992acceleration}.
As the teacher generally outperforms the student model, we treat the images generated by the teacher model as pseudo labels. 

\begin{figure*}[htbp]
	\centering
	\includegraphics[width=1.0\textwidth]{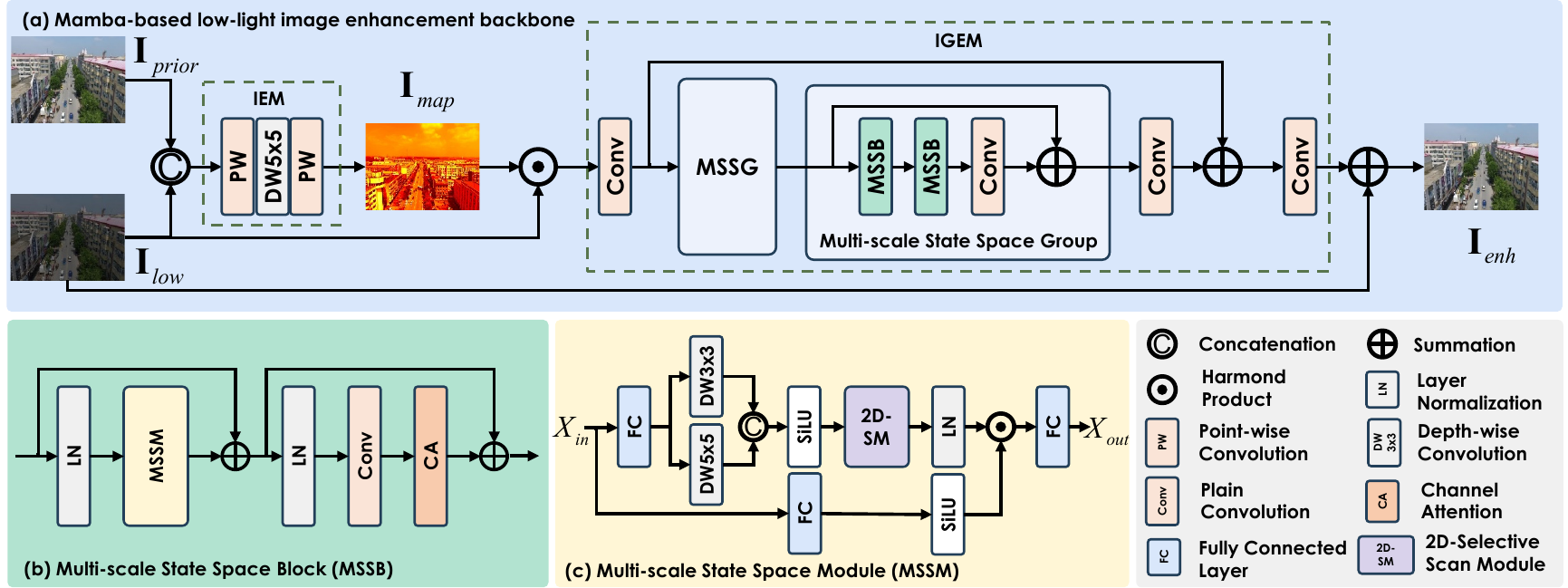}
	\caption[]{The pipeline of our Mamba-based low-light image enhancement backbone (a), which mainly consists of an illumination estimation module (IEM) and an illumination-guided enhancement module (IGEM). The IGEM processes two building blocks, the multi-scale state space block (MSSB) (b) and the multi-scale state space module (MSSM) (c). }
	\vspace{-0.2in}
	\label{fig:multiscale_mamba}
\end{figure*}

\subsection{Mamba-based Low-light Image Enhancement Backbone}
Mamba-based restoration network MambaIR~\cite{guo2024mambair} can balance long-range pixel dependencies establishment and computational efficiency. Besides, it has demonstrated significant effectiveness and achieved impressive success in image restoration due to its inherent capability of activating more pixels.
Despite its success, the MambaIR~\cite{guo2024mambair} has limitations when applied directly to corporate with our Semi-LLIE framework. 
Although MambaIR~\cite{guo2024mambair} has strong long-range pixel modeling capabilities, its insufficient focus on local pixel relationships leads to detail loss in the enhanced images.
We design a specific Mamba-based low-light image enhancement backbone to overcome this limitation, focusing on establishing robust global-local pixel relationships. As depicted in Fig.~\ref{fig:multiscale_mamba} (a), our Mamba-based low-light image enhancement backbone primarily consists of two stages: illumination estimation module and illumination-guided enhancement module.

Given a low-light input image $I_{low} \in \mathbb{R}^{H \times W \times 3}$, 
We first compute the mean of $I_{low}$ along the channel dimension to obtain the brightness prior $I_{prior} \in \mathbb{R}^{H \times W \times 1}$.
We then concatenate $I_{low}$ and $I_{prior}$ and employ an illumination estimation module to generate the illumination map $I_{map} \in \mathbb{R}^{H \times W \times C}$, where $H$, $W$ and $C$ denotes the height, width, and channel counts. The illumination estimation module is constructed by three convolution layers, a point-wise convolution layer to expand the channel dimension to $C$, a depth-wise convolution layer, and another point-wise convolution layer to map the channel dimension back to $3$. 
Subsequently, low-light input image $I_{low}$ and the illumination map $I_{map}$ are concatenated and fed into the illumination-guided enhancement module.
The enhancement module first utilizes a plain convolution layer to extract shallow features. Then two cascaded multi-scale state-space groups (MSSGs) and another plain convolution layer are applied to generate the lighted-up deep features. Moreover, the shallow and light-up deep features are fused by a pixel-wise summation, following an additional plain convolution layer to generate a residual image $I_{res}$.
Finally, we obtain the enhanced image by combining the low-light input image and the residual image: $I_{enh} = I_{low} + I_{res}$. 
Our Mamba-based low-light image enhancement backbone introduces several key modifications based on the MambaIR~\cite{guo2024mambair} model. Firstly, we design an illumination estimation module to generate the illumination map which is then used to guide the image enhancement process. Secondly, we introduce a multi-scale state-space block (MSSB) to enhance the local representation ability of the original vision state-space block (VSSB) in MambaIR~\cite{guo2024mambair}. Thirdly, we employ both feature-level and image-level residual learning strategies.

\subsubsection{Multi-scale State-Space Block}
Although vision state-space block (VSSB) in MambaIR~\cite{guo2024mambair} can model global pixel dependencies, we find that simply applying VSSB to solve the image enhancement problem yields only suboptimal outcomes. 
Hence, designing a novel block specifically for a Mamba-based low-light image enhancement backbone holds considerable potential.
In response, we propose the multi-scale state-space block (MSSB), which adapts the multi-scale state-space module (MSSM) for the image enhancement task.
As shown in Fig.~\ref{fig:multiscale_mamba} (b), we largely followed the design choices of the transformer-style VSSB. However, our MSSB differs from the original VSSB in one key aspect. We have developed MSSM to replace VSSM, aiming to extract more robust global-local feature representations from low-light input images. 

\subsubsection{Multi-scale State-Space Module}
Inspired by Mamba's success in modeling long-range pixel relationships with linear efficiency, the authors of MambaIR~\cite{guo2024mambair} have introduced the vision state-space module (VSSM) to image restoration. 
The VSSM captures long-range dependencies using the state space equation. However, it does not specifically focus on modeling dependencies between local region pixels. It typically integrates single-scale depth-wise convolutions and overlooks the interdependencies among multi-scale low-light regions.
This issue can lead to occasional detail loss in the restoration results and is particularly pronounced when applied to low-light enhancement tasks. Certainly, the efficacy of multi-scale feature representations of images has been well established in restoring degraded images~\cite{chen2023learning}.
Therefore, we design MSSM to enhance the VSSM by incorporating a multi-scale feature learning mechanism to enhance local feature representation. The structure of MSSM is shown in Fig.~\ref{fig:multiscale_mamba} (c).

Following~\cite{liu2024vmamba, guo2024mambair}, the input tensor $X_{in} \in \mathbb{R}^{H \times W \times C}$ of MSSM is processed through two concurrent branches. 
In the upper branch, we first expand the channel number of the input feature to $\lambda C$ by a fully connected layer, where $\lambda$ is a fixed channel expansion factor.
Then, we set up two parallel depth-wise convolutions for each feature block with different kernel sizes, $3 \times 3$ and $5 \times 5$, respectively. 
Following this way, we can produce robust multi-scale local features benefiting detail preserving. 
The multi-scale local features are then fused by concatenation for further processing.
After that, a SiLU~\cite{shazeer2020glu} activation function, a 2D-SSM~\cite{guo2024mambair} layer, and a layer normalization are operated sequentially to refine the multi-scale features. We apply the 2D-SSM layer introduced in~\cite{guo2024mambair} without any changes.
In the bottom branch, the channel number of the features is increased to $\lambda C$ utilizing a fully connected layer succeeded by a SiLU~\cite{shazeer2020glu} activation function. 
Subsequently, the output features generated by the two branches are fused using the Hadamard product operation. Finally, we utilize another fully connected layer to project back the channel number of features to $C$ and then obtain the output feature $X_{out}$. The whole computational process of MSSM can be formulated as:

\begin{equation}
    \begin{aligned}
        &X_{0} = \mathrm{FC}(X_{in}), \\ 
        &X_{1} = \mathrm{DConv3}(X_{0}), \\
        &X_{2} = \mathrm{DConv5}(X_{0}), \\
        &X_3 = \mathrm{LN(2D\text{-}SSM(SiLU}(\left[X_{1},X_{2})\right]))),\\
        &X_4 = \mathrm{SiLU(FC}(X_{in})),\\
        &X_{out} = \mathrm{FC}(X_3 \odot X_4),
    \end{aligned}
\end{equation}
where $\mathrm{FC}$ denotes the fully connected layer. $\mathrm{DConv3}$ and $\mathrm{DConv5}$ denotes depth-wise convolution layer with kernel size $3 \times 3$ and $5 \times 5$. $\mathrm{LN}$ and $\left[ \cdot \right]$ represent the layer normalization and concatenation operation, respectively. The $\odot$ represents the Hadamard product.

\subsection{Semantic-Aware Contrastive Loss}
Methods with mean teacher structure usually employ $L_1$ loss to keep consistency between the teacher and the student model. However, the application of $L_1$ consistency loss can lead to the issue of overfitting the student model to incorrect predictions. We propose to mitigate this issue by incorporating contrastive loss in the optimization process. Contrastive learning has been recently adapted to tackle image restoration tasks~\cite{liang2022semantically, wu2021contrastive} and achieved significant success. In these works, the contrastive loss is constructed using paired datasets, wherein positive samples are aligned with the ground truth and negative samples are associated with the deteriorated images. In contrast, we intend to integrate contrastive loss into the analysis of unpaired data. We employ a distinct method to construct positive and negative sample pairs to achieve this. 
We treat the teacher's $\hat{I}_{t}^u$ and the student's output $\hat{I}_{s}^u$ as positive and anchor samples, respectively. The original input low-light image $I^u$ is viewed as a negative sample. The semantic-aware contrastive loss is calculated as follows:
\begin{equation}\label{eq4}
L_{scr}=\omega\frac{||\mathcal{E}(\hat{I}_{s}^u),\mathcal{E}(I^u)||_1}{||\mathcal{E}(\hat{I}_{t}^u),\mathcal{E}(I^u)||_1},
\end{equation}
\noindent
where $\hat{I}_{s}^u= f_s(I_{s}^u)$ represents the student's output on the unpaired strongly augmented low-light input $I_{s}^u$. $\hat{I}_{t}^u= f_t(I_{t}^u)$ stands for the teacher's prediction on the weakly augmented input $I_{t}^u$. $\mathcal{E}(\cdot)$ represents the output embedding generated by the image encoder of RAM~\cite{zhang2024recognize}. $\omega$ denotes the weighting factor. $L_1$ loss is applied to evaluate the distance between the enhanced image of the student model with the negative and positive samples in RAM's image embedding space. 

The RAM~\cite{zhang2024recognize} adopts a Swin-Transformer~\cite{liu2021swin} to implement the image encoder, which is structured into four stages, each progressively decreasing the resolution. Different stages are constructed by varied numbers of Swin-Transformer blocks~\cite{liu2021swin}. The downsampling across distinct stages is implemented by the path merging operation~\cite{liu2021swin}. We choose RAM's image encoder~\cite{zhang2024recognize} because it is trained on large-scale and refined text-image pairs, thus providing us with more informative and semantically meaningful features. Based on this, our semantic-aware contrastive loss can preserve semantic consistency between the teacher and student models. This contributes significantly to mitigating color cast and producing visually appealing enhanced images with natural colors. 

\subsection{RAM-based Perceptual Loss}\label{sec:sam_encoder}
To restore realistic texture details in the enhanced images, we design a novel perceptual loss function based on the intermediate features extracted from the last three stages of the RAM's pre-trained image encoder~\cite{zhang2024recognize}. Built on the ideas of VGG-based perceptual loss~\cite{ledig2017photo}, our RAM-based perceptual loss is also designed to evaluate the perceptual similarity between the two input images. Besides, 
due to being trained on large-scale text-image pairs, the intermediate features of RAM's image encoder exhibit more semantically meaningful information. Moreover, RAM demonstrates exceptional generalization ability thus it can effectively mitigate the impact of low-level local color or texture discrepancies between two compared images.
Above, our RAM-based perceptual loss is more appropriate for evaluating the semantic similarity of images captured in diverse scenes compared with the VGG-based perceptual loss~\cite{ledig2017photo}.

Let $\mathcal{E}_i$ denote the output features extracted from the $i$-th stage within the RAM's image encoder as previously described.
We define the RAM-based perceptual loss function to be the distance between the output feature representations of an enhanced image $I^{enh}$ and the ground truth image $I^l_{gt}$. It can be formulated as follows:
\begin{equation}
    \begin{split}
        \mathcal{L}_{ramper} =
        \frac{1}{W_{i}H_{i}} & \sum_{x=1}^{W_{i}} \sum_{y=1}^{H_{i}} (\mathcal{E}_i(I^l_{gt})_{x,y} - \mathcal{E}_i(I^{enh})_{x,y})^2.
    \end{split}
    \label{eq:sam_perceptual}
\end{equation}
Here $W_{i}$ and $H_{i}$ denote the spatial dimensions of the features extracted at the $i$-th stage within the RAM's image encoder.

\subsection{Overall Objective} \label{sec:overall_obj}

The overall objective is composed of a supervised part, an unsupervised part, and an adversarial part. For the supervised part, we incorporate the fidelity loss $\mathcal{L}_{sup}$ introduced in Section~\ref{sec:overall_structure}, a gradient loss $\mathcal{L}_{gradient}$ following~\cite{huo2021efficient, huang2023contrastive}, and our RAM-based perceptual loss $\mathcal{L}_{ramper}$. The $\mathcal{L}_{gradient}$ and $\mathcal{L}_{ramper}$ are utilized to maintain the textural details of the enhanced images. The gradient loss is as follows:
\begin{equation}\label{eq:loss_sup_grad}
    \mathcal{L}_{grad} = ||f_g(I_{enh}) - f_g(I_{gt}^l)||_1,
\end{equation}
where $f_g$ denotes the function to generate a gradient map~\cite{ma2021structure} and $|.|$ represents the $L_1$ distance. 
Above all, the supervised objective can be formulated as follows:
\begin{equation}\label{eq:loss_sup_all}
    \mathcal{L}'_{sup} = \mathcal{L}_{sup} + \gamma_1\mathcal{L}_{ramper} + \gamma_2 \mathcal{L}_{grad}.
\end{equation}

For the unsupervised part, we propose integrating the semantic-aware contrastive loss $\mathcal{L}_{scr}$ with the original $\mathcal{L}_{un}$ consistency loss introduced in Section~\ref{sec:overall_structure}. The unsupervised objective can be represented as:
\begin{equation}\label{eq:loss_unsup_all}
    \mathcal{L}'_{un} = \mathcal{L}_{un} + \mathcal{L}_{scr}.
\end{equation}

Besides, we employ a PatchGAN~\cite{isola2017image} discriminator to differentiate between the generated outputs and real images as shown in Fig.~\ref{fig:overall_framework}. It facilitates generating faithful content and promoting image-level authenticity in terms of color and textures. 
For the enhancement operation $\textit{G}_{H}(\cdot)$: $I_{low}$ $\rightarrow$ $I_{enh}$, we denote the discriminator as $\textit{D}_{H}$, which facilitates to learn a mapping between the low-light domain $\boldsymbol{\mathcal{L}}$ and normal-light domain $\boldsymbol{\mathcal{H}}$. 
The adversarial objective function can be formulated as follows:
\begin{equation}
\begin{aligned}
	\label{shi3}
	\mathcal{L}_{adv} &= \mathbb{E}_{I_{enh} \sim \boldsymbol{\mathcal{H}}}[\log\textit{D}_{H}(I_{enh})] \\
	&+ \mathbb{E}_{I_{low} \sim \boldsymbol{\mathcal{L}}}[\log(1-\textit{D}_{H}(\textit{G}_{H}(I_{low}))],
\end{aligned}
\end{equation}
where $\textit{D}_{H}$ aims to determine whether an image is enhanced (fake) or captured (real), distinguishing the enhanced images $\textit{G}_{H}(I_{low})$ from the real normal-light domain $\boldsymbol{\mathcal{H}}$. Meanwhile, $\textit{G}_{H}(\cdot)$ strives to deceive $\textit{D}_{H}$ by producing results that are indistinguishable from  $\boldsymbol{\mathcal{H}}$. 
During training, our Mamba-based low-light image enhancement backbone is viewed as the generator and tries to minimize the adversarial objective, while the discriminator aims to maximize it. 

The overall objective to train our Semi-LLIE is formulated as follows:
\begin{equation}\label{eq:loss_all}
    \mathcal{L}_{overall} = \mathcal{L}_{sup}'+ \lambda_1 \mathcal{L}_{un}' + \lambda_2 \mathcal{L}_{adv},
\end{equation}
where $\lambda_1$ and $\lambda_2$ are the scalar factors to balance varied parts.


\section{Experiments}

\subsection{Implementation Details}
Our Semi-LLIE is implemented using the PyTorch library~\cite{paszke2019pytorch}. We perform all experiments on two NVIDIA Tesla A100 GPUs. For training, we utilize the AdamP~\cite{adamp} optimizer with parameters $\beta_1 = 0.9$, $\beta_2 = 0.999$, and the weight decay value $1e^{-4}$.
Our Semi-LLIE is trained for $200$ epochs in total. The learning rate is set as $2e^{-4}$ for the first $100$ epochs and is linearly decreased to $0$ in the last $100$ epochs.
The batch size is set to $16$, where $8$ of them are paired image pairs and others are unpaired low-light images. We apply random crop and rotation on the paired training data, where the images are resized to $256 \times 256$. We augment the unpaired data with Gaussian blur, grayscale conversion, and color jittering operations to feed the student model. The unpaired images forwarded to the teacher model are only resized to match the input resolutions.
The scaled balance factors of different loss terms are set as follows:
$\gamma_1 = 0.1$, $\gamma_2 = 0.1$, and $\lambda_1$ is adjusted based on a warming up function~\cite{liu2021synthetic, huang2023contrastive}: $\lambda_1 (t)=0.2\times e^{-5(1-t/200)^2}$, where $t$ refers to training epochs. $\lambda_2$ is fixed at $0.1$ empirically.

\begin{table*}[ht]
	\centering
    \caption{Quantitative Comparison on one non-referenced dataset Visdrone~\cite{zhu2021detection} and one full-referenced datasets LSRW~\cite{hai2023r2rnet}. The top result and the second-best result are marked in \textbf{bold} and \underline{underlined}, respectively.}

	\resizebox{\linewidth}{!}{
	\begin{tabular}{c|c|cccccccc|cccccc|cc}
		\toprule[1.5pt]
		\multirow{2}{*}{Datasets} & \multirow{2}{*}{Metrics}  & \multicolumn{8}{c|}{\cellcolor{gray!40}\emph{Supervised}} & \multicolumn{6}{c|}{\cellcolor{gray!40}\emph{Unsupervised}} & \multicolumn{2}{c}{\cellcolor{gray!40}\emph{Semi-supervised}} \\ \cline{3-18}
		& & \cellcolor{gray!40}RetinexNet & \cellcolor{gray!40}MIRNet & \cellcolor{gray!40}URetinex & \cellcolor{gray!40}Restormer & \cellcolor{gray!40}SwinIR & \cellcolor{gray!40}SNR-Net & \cellcolor{gray!40}RetinexFormer & \cellcolor{gray!40}MambaIR & \cellcolor{gray!40}SSIENet & \cellcolor{gray!40}RUAS & \cellcolor{gray!40}EnGAN & \cellcolor{gray!40}ZeroDCE & \cellcolor{gray!40}SCI & \cellcolor{gray!40}NeRCo & \cellcolor{gray!40}DRBN & \cellcolor{gray!40}Ours \\ 
  
        \midrule[1.5pt]
        
  	\multirow{5}{*}{Visdrone~\cite{zhu2021detection}}
  		& FID $\downarrow$ & 146.19 & 39.68 & 53.98 & 40.67 & 84.10 & 71.39 & 42.54 & \underline{38.32} & 67.53 & 49.74 & 81.71 & 62.14 & 84.16 & 49.87 & 75.77 & \textbf{36.49}\\
		& NIQE $\downarrow$ & 4.077 & 3.784 & 3.743 & 3.789 & 3.970 & 7.150 & 3.729 & 3.965 & 3.881 & 3.769 & 3.855 & 3.740 & 3.921 & \underline{3.725} & 3.851 & \textbf{3.667}\\
  		& LOE $\downarrow$ & 508.7 & 235.8 & \underline{222.3} & 225.3 & 246.8 & 508.7 & 251.7 & 238.5 & 363.2 & 335.2 & 249.2 & 246.0 & 267.1 & 240.8 & 232.6 & \textbf{204.1}\\

        \midrule


		
        \multirow{4}{*}{LSRW~\cite{hai2023r2rnet}} & PSNR $\uparrow$ & 15.48 & 19.27 & 18.10 & 19.11 & 19.04 & 18.98 & \underline{19.36} & 19.34 & 16.14 & 14.11 & 17.06 & 15.80 & 15.24 & 19.00 & 17.56 & \textbf{19.73} \\
		& SSIM $\uparrow$ & 0.3468 & 0.5650 & 0.5149 & 0.5813 & 0.5678 & 0.5639 & \underline{0.5872} & 0.5871 & 0.4627 & 0.4112 & 0.4601 & 0.4450 & 0.4192 & 0.5360 & 0.5626 & \textbf{0.7840} \\ \cline{2-18}
		& NIQE $\downarrow$ & 18.312 & 5.334 & 11.681 & 5.189 & \underline{4.825} & 4.896 & 5.489 & 4.395 & 12.702 & 11.085 & 11.946  & 11.833 & 10.227 & 9.232 & 5.751 & \textbf{3.974}\\
		& LOE $\downarrow$ & 346.3 & 232.7 & 218.5 & 243.6 & 265.7 & 279.1 & 205.7 & 210.37 & 196.0 & 198.9 & 385.1 & 216.0 & 234.6 & \underline{198.4} & 236.5 & \textbf{195.0}\\ 

        \midrule

        \multirow{2}{*}{Complexity}
  		& Params (M) $\downarrow$ & 0.84 & 31.76 & 10.37 & 26.13 & 0.90 & 4.01 & 1.61 & 3.01 & 0.68 & \underline{0.003} & 114.35 & 0.079 & \textbf{0.0003} & 23.29 & 5.27 & 0.46 \\
		& FLOPS (G) $\downarrow$ & 587.47 & 785.00 & 34.61 & 144.25 & 64.44 & 26.35 & 15.57 & 170.34 & 34.60 & \underline{0.83} & 61.01 & 5.21 & \textbf{0.0619} & 177.74 & 48.61 & 26.21 \\
                
        \bottomrule[1.5pt]
	\end{tabular}
	}
        \label{tab:full_referenced_comp_results}
        \vspace{-1.0em}

\end{table*}

\subsection{Datasets and Metrics}

To verify the efficacy of our Semi-LLIE, We train and test our model on the VisDrone dataset~\cite{zhu2021detection} and the LSRW dataset~\cite{hai2023r2rnet}. The VisDrone dataset~\cite{zhu2021detection} is recorded by unmanned aerial vehicles and annotated for multiple tasks. Leveraging this dataset, we developed a novel dataset tailored for low-light enhancement tasks. We selected $5746$ clear daytime images and $1915$ nighttime images for our experiment. Our training set is composed of a paired sub-dataset and an unpaired sub-dataset. The paired sub-dataset contains $5213$ clear daytime images and corresponding synthesized low-light images. The unpaired sub-dataset contains $1258$ low-light drone images without normal-light images. The remaining $657$ unpaired low-light drone images are for testing. For validation, we select $533$ normal-light images with paired synthesized low-light images. We have summarized the comprehensive statistics of the VisDrone dataset in Table~\ref{tab:semi-dataset-visdrone}.

\setlength{\tabcolsep}{4pt}
\begin{table}[ht]
	\centering
 	\caption{The numbers of training, validation, and testing images from the tailored Visdrone dataset~\cite{zhu2021detection}.}
	\begin{tabular}{c|c|c|c}
		\toprule[1.5pt]
		\cellcolor{gray!40}Sub-dataset & \cellcolor{gray!40}normal-light & \cellcolor{gray!40}low-light (syn) & \cellcolor{gray!40}low-light (real) \\
        \midrule[1.5pt]
        paired (for train) & 5213 & 5213 & -\\
		\hline
		unpaired (for train) & - & - & 1258 \\
		\hline
		val & 533 & 533 & - \\
		\hline
		unpaired (for test) & - & - & 657 \\  
		\hline
		
		Total & 5746 & 5746 & 1915\\
		\bottomrule[1.5pt]
	\end{tabular}
	\label{tab:semi-dataset-visdrone}
\end{table}

To provide a more comprehensive and convincing comparison, we have extended our evaluation to a full-referenced LSRW dataset~\cite{hai2023r2rnet} captured at real-world scenes.
The LSRW dataset~\cite{hai2023r2rnet} consists of $5650$ paired images, where $5600$ pairs are selected for training and the rest $50$ for testing. Each pair contains an under-exposed image and a ground truth image captured in real-world scenarios. 
To train our Semi-LLIE, we have to construct a paired sub-dataset and an unpaired sub-dataset. For the paired dataset, we follow~\cite{ma2022sci, yang2023implicit} and apply the LSRW~\cite{hai2023r2rnet} dataset.
For the unpaired dataset, we utilize the dataset proposed in EnGAN~\cite{jiang2021enlightengan}, which contains $914$ low-light images and $1016$ normal-light images. Note that, we only use the low-light images for training to validate the superiority of our semi-supervised learning strategy.

We utilize commonly used full-referenced metrics PSNR and SSIM~\cite{wang2004ssim} to evaluate the performances on paired datasets while adopting the no-reference metrics FID~\cite{heusel2017gans}, NIQE~\cite{mittal2012niqe} and LOE~\cite{wang2013npe} to evaluate different comparison methods on the unpaired datasets. A lower FID~\cite{heusel2017gans}, NIQE~\cite{mittal2012niqe}, or LOE~\cite{wang2013npe} represents a more perceptually favored quality.


\subsection{Comparison with Existing Methods}

We compare our Semi-LLIE with several representative low-light image enhancement methods, including eight supervised methods (RetinexNet~\cite{Chen2018Retinex}, MIRNet~\cite{zamir2020mirnet}, URetinex-Net~\cite{wu2022uretinexnet}, Restormer~\cite{zamir2022restormer}, SwinIR~\cite{liang2021swinir}, RetinexFormer~\cite{cai2023retinexformer}, and MambaIR~\cite{guo2024mambair}), six unsupervised methods (SSIENet~\cite{zhang2020self}, RUAS~\cite{liu2021ruas}, EnGAN~\cite{jiang2021enlightengan}, ZeroDCE~\cite{guo2020zerodce}, SCI~\cite{ma2022sci}, and NeRCo~\cite{yang2023implicit}) and one previously developed semi-supervised method (DRBN~\cite{yang2020drbn}). Note that, all the comparison methods are trained from scratch on the constructed Visdrone~\cite{zhu2021detection} or LSRW~\cite{hai2023r2rnet} datasets.

\subsubsection{Quantitative Results}
As depicted in Table~\ref{tab:full_referenced_comp_results}, our Semi-LLIE demonstrates SOTA performance across various benchmarks, as verified by both no-reference and full-reference metrics. This highlights the effectiveness of our proposed Semi-LLIE framework. 

On the non-referenced Visdrone~\cite{zhu2021detection} dataset, our Semi-LLIE significantly outperforms the comparison methods in FID~\cite{heusel2017gans}, NIQE~\cite{mittal2012niqe}, and LOE~\cite{wang2013npe} metrics. Notably, our method even outperforms some supervised approaches. It shows that our method recovers more visual-friendly images than comparison methods.
Specifically, compared with recent competitive unsupervised methods like SCI~\cite{ma2022sci} and NeCRo~\cite{yang2023implicit}, our method introduces semantic feature alignment by the proposed semantic-aware contrastive loss. It enables us to generate perceptual-friendly enhanced images with natural colors in real-world scenes.

On the full-referenced LSRW~\cite{zhu2021detection} datasets, our Semi-LLIE performs the best in all evaluation metrics. Our Semi-LLIE outperforms the SOTA-supervised method Retinexformer~\cite{cai2023retinexformer} by $0.37$ in PSNR and $0.1968$ in SSIM. Compared with the previous SOTA unsupervised method NeRCo, we obtain a $0.73$ dB gain in PSNR and $0.2480$ in SSIM. 
Compared with the previous semi-supervised method DRBN~\cite{yang2020drbn}, we obtain a $2.17$ dB gain in PSNR and $0.2214$ in SSIM. 
The higher SSIM results of our Semi-LLIE demonstrate its superiority in restoring structural details and natural colors, which are further validated by the subsequent qualitative experimental results. This contributes to the involvement of the semantic-aware contrastive loss, RAM-based perceptual loss, and our Mamba-based low-light image enhancement backbone. 
Apart from the superiority in full-referenced metrics, we also achieve favorable results in non-referenced metrics NIQE and LOE. This suggests that our Semi-LLIE framework supports us in producing user-friendly enhanced results on the LSRW dataset~\cite{hai2023r2rnet} too.
All the results indicate that our Semi-LLIE is better suited for handling real-world scenes and can more effectively utilize unpaired and paired training data simultaneously.

We analyze the complexity of the comparison methods using an input resolution of $256 \times 256$ and present the results in Table~\ref{tab:full_referenced_comp_results}. Compared with Transformer-based supervised methods, Semi-LLIE consumes only $18.16\%$ and $40.67\%$ GFLOPS, and $1.76\%$ and $51.11\%$ Params of Restormer~\cite{zamir2022restormer} and SwinIR~\cite{liang2021swinir}. Compared with the supervised Mamba-based method MambaIR~\cite{guo2024mambair}, our Semi-LLIE costs $15.28\%$ Params and only $15.38\%$ GFLOPS. Compared with the current SOTA unsupervised method NeRCo~\cite{yang2023implicit}, our Semi-LLIE only requires merely $1.97\%$ Params and $14.74\%$ GFLOPS. To summarize, our Semi-LLIE achieves performance comparable to current SOTA methods with moderate computational cost. These results highlight the efficacy and efficiency of the proposed Semi-LLIE.

\begin{figure*}[t]
  \centering
  \begin{subfigure}[c]{0.160\linewidth}
    \includegraphics[width=1.\linewidth]{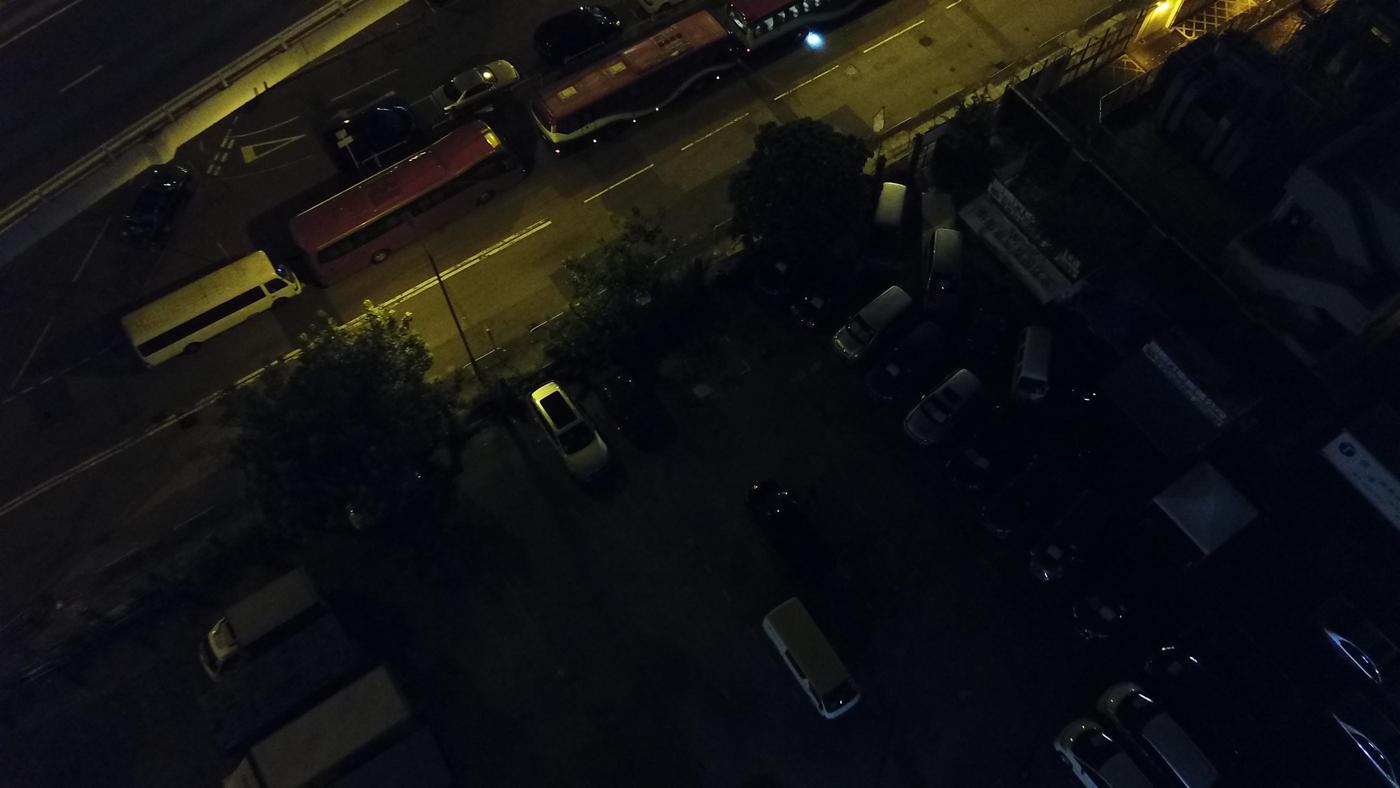}\vspace{-0.2em}
    \caption*{(a) Input}\medskip
    
    \includegraphics[width=1.\linewidth]{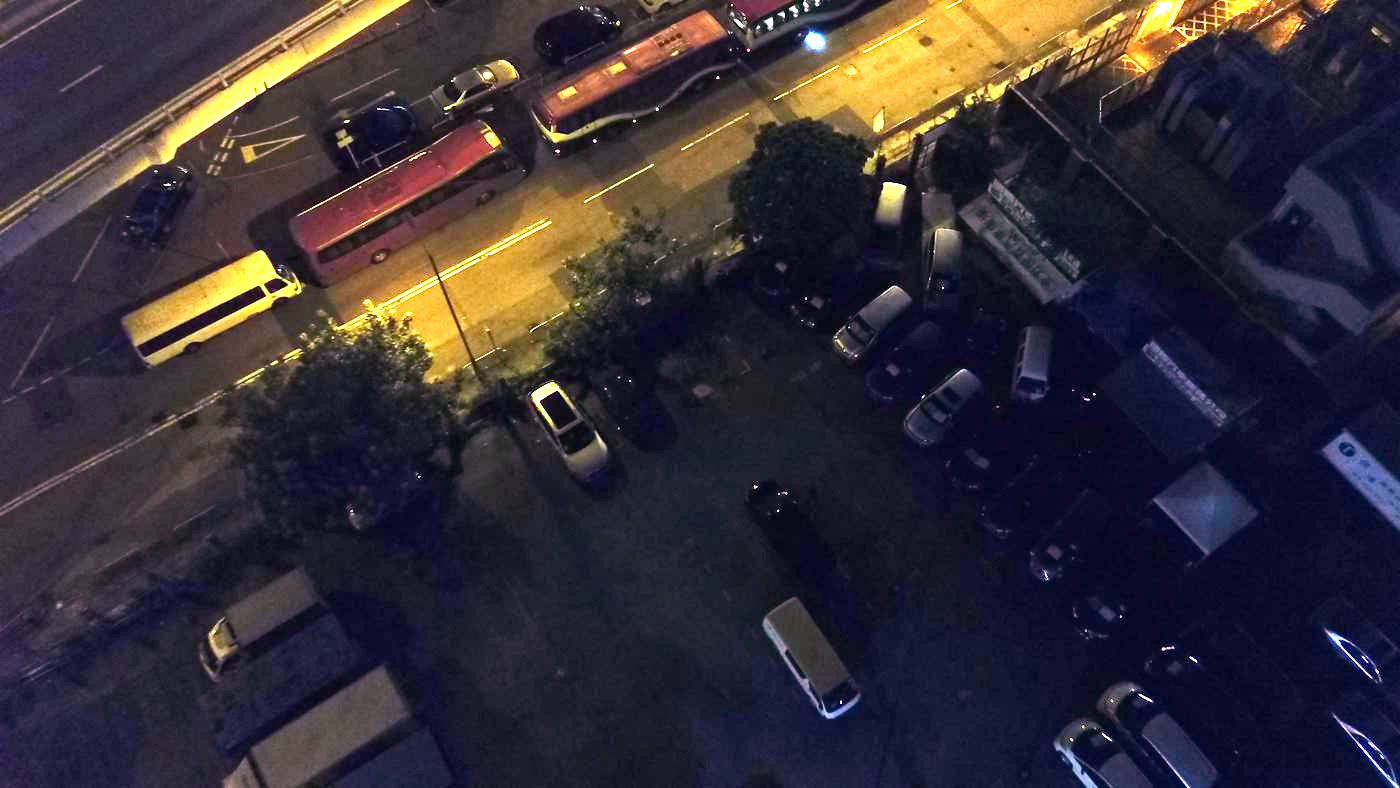}\vspace{-0.2em}
    \caption*{(g) RUAS~\cite{liu2021ruas} }\medskip
  \end{subfigure}
  \hfill
  \begin{subfigure}[c]{0.160\linewidth}
    \includegraphics[width=1.\linewidth]{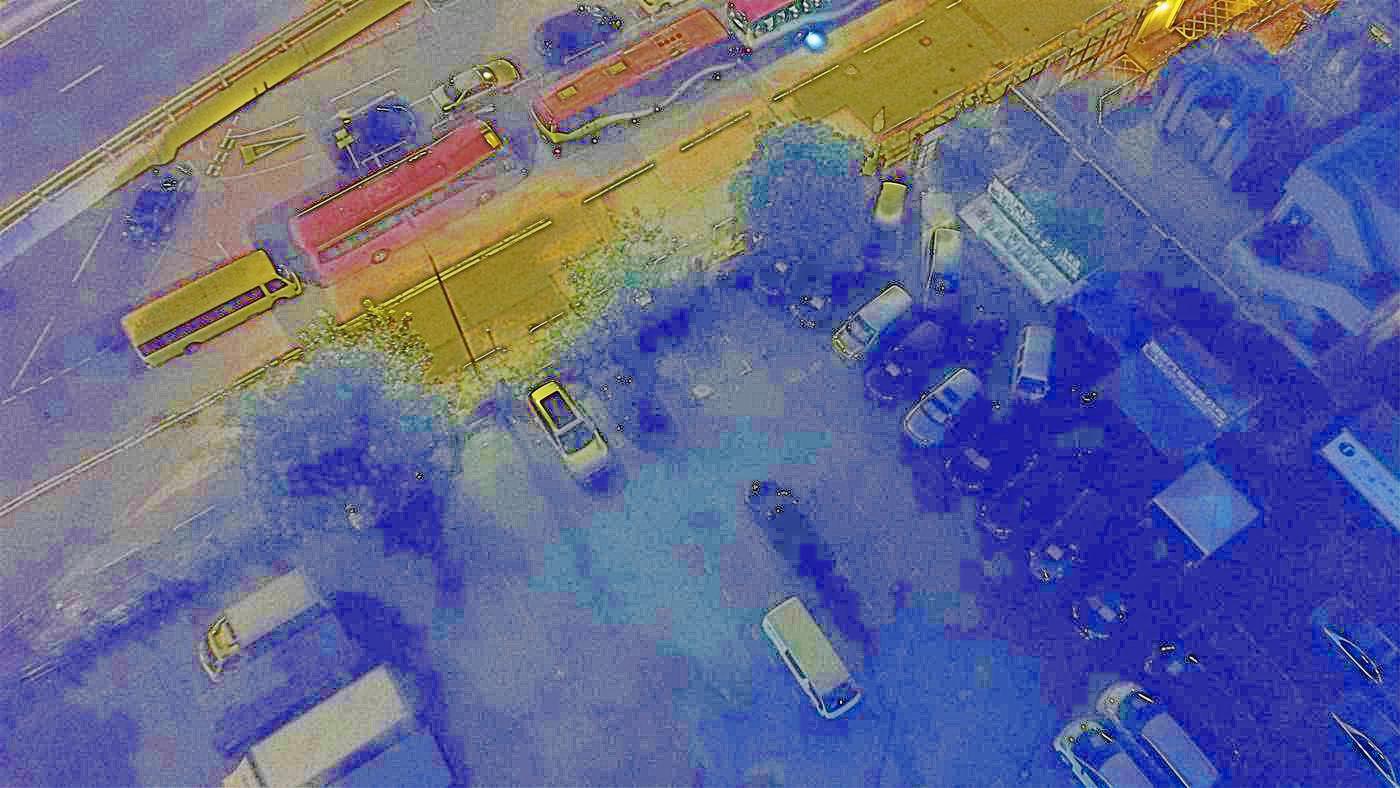}\vspace{-0.2em}
    \caption*{(b) RetinexNet~\cite{Chen2018Retinex}}\medskip
    
    \includegraphics[width=1.\linewidth]{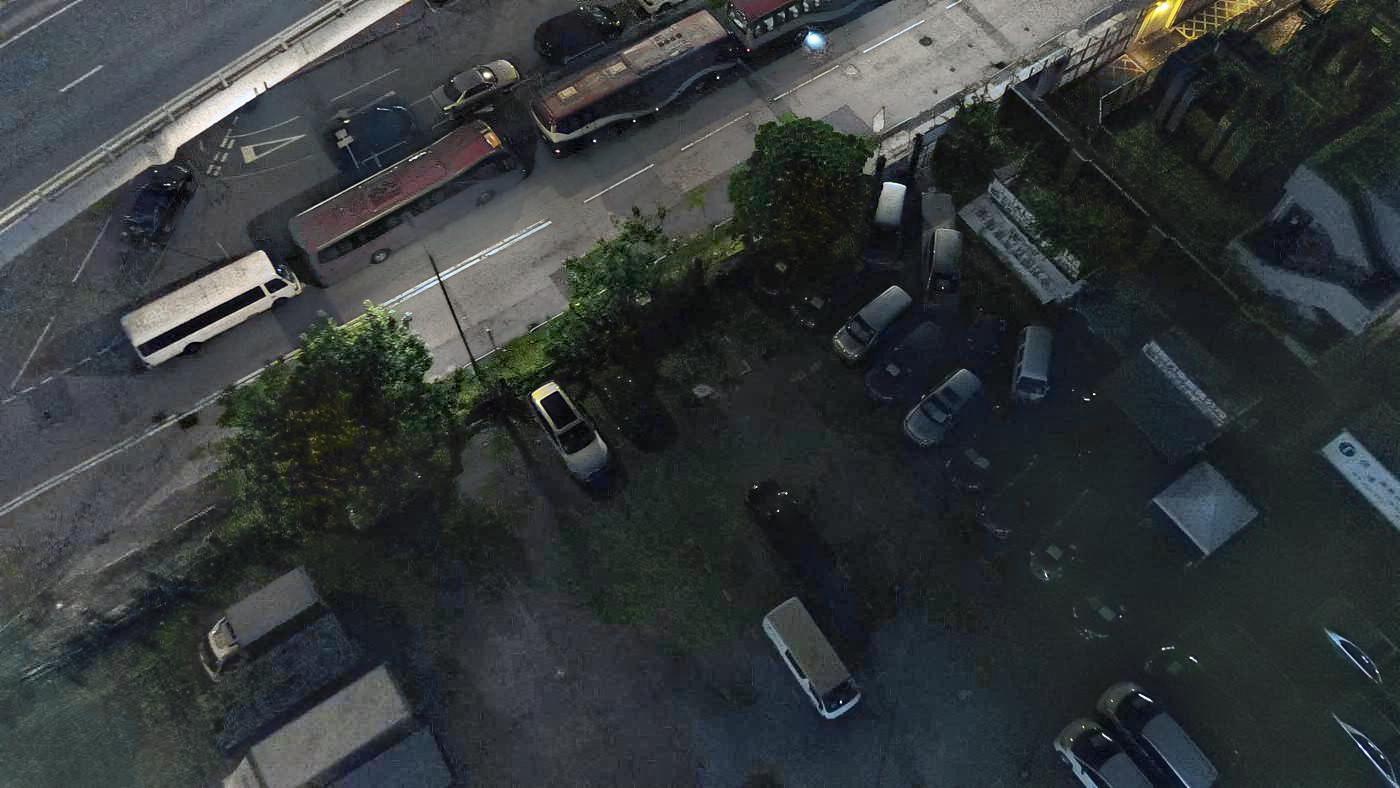}\vspace{-0.2em}
    \caption*{(h) EnGAN~\cite{jiang2021enlightengan}}\medskip
  \end{subfigure}
  \hfill
  \begin{subfigure}[c]{0.160\linewidth}
    \includegraphics[width=1.\linewidth]{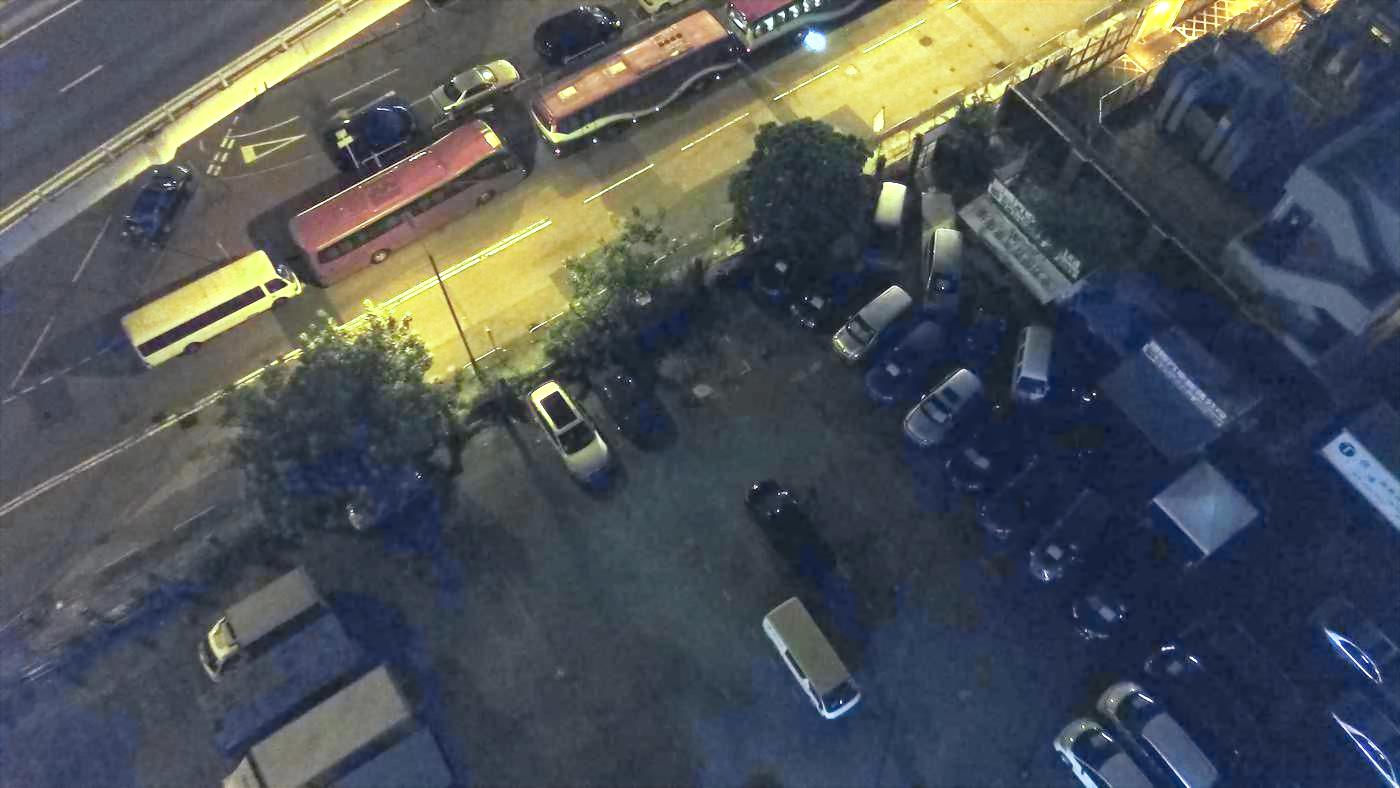}\vspace{-0.2em}
    \caption*{(c) MIRNet~\cite{zamir2020mirnet}}\medskip
    
    \includegraphics[width=1.\textwidth]{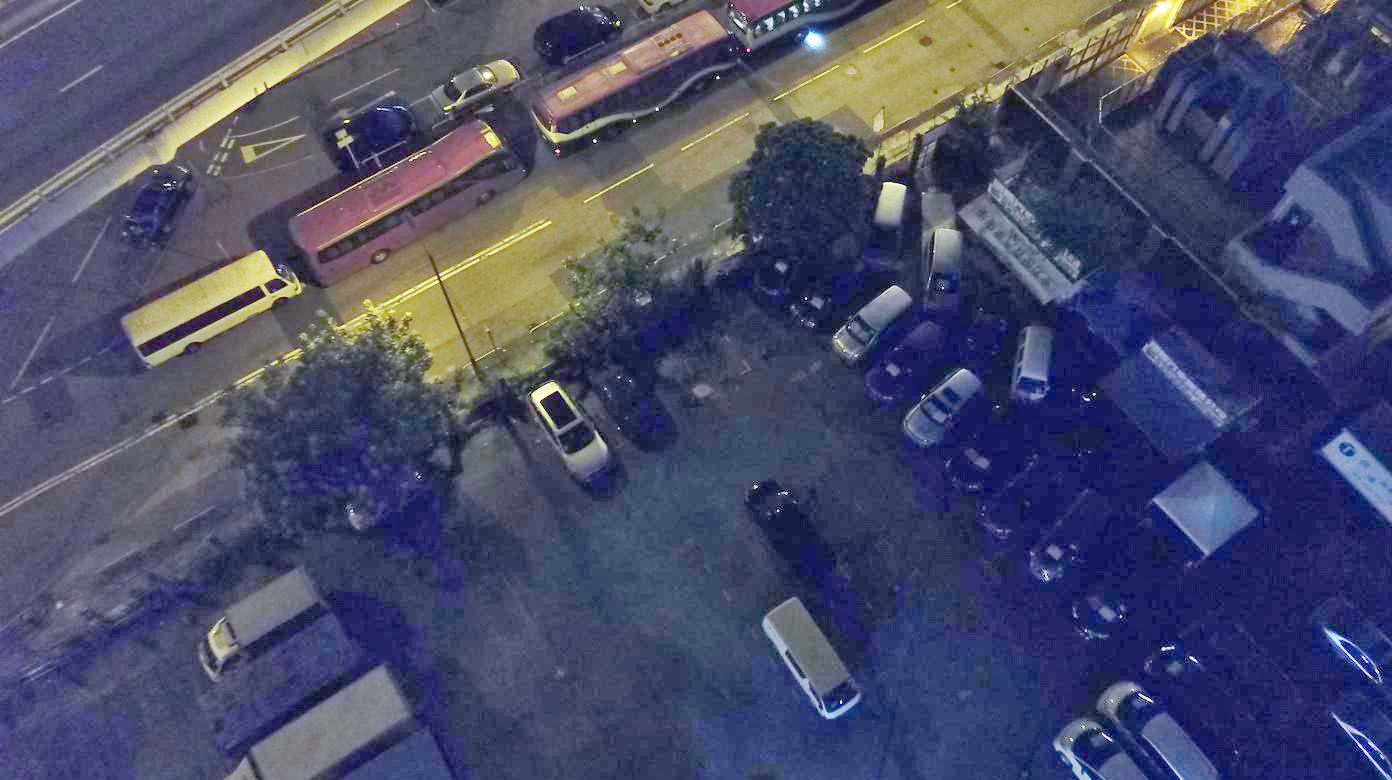}\vspace{-0.2em}
    \caption*{(i) ZeroDCE~\cite{guo2020zerodce}}\medskip
  \end{subfigure}
  \hfill
  \begin{subfigure}[c]{0.160\textwidth}
    \includegraphics[width=1.\textwidth]{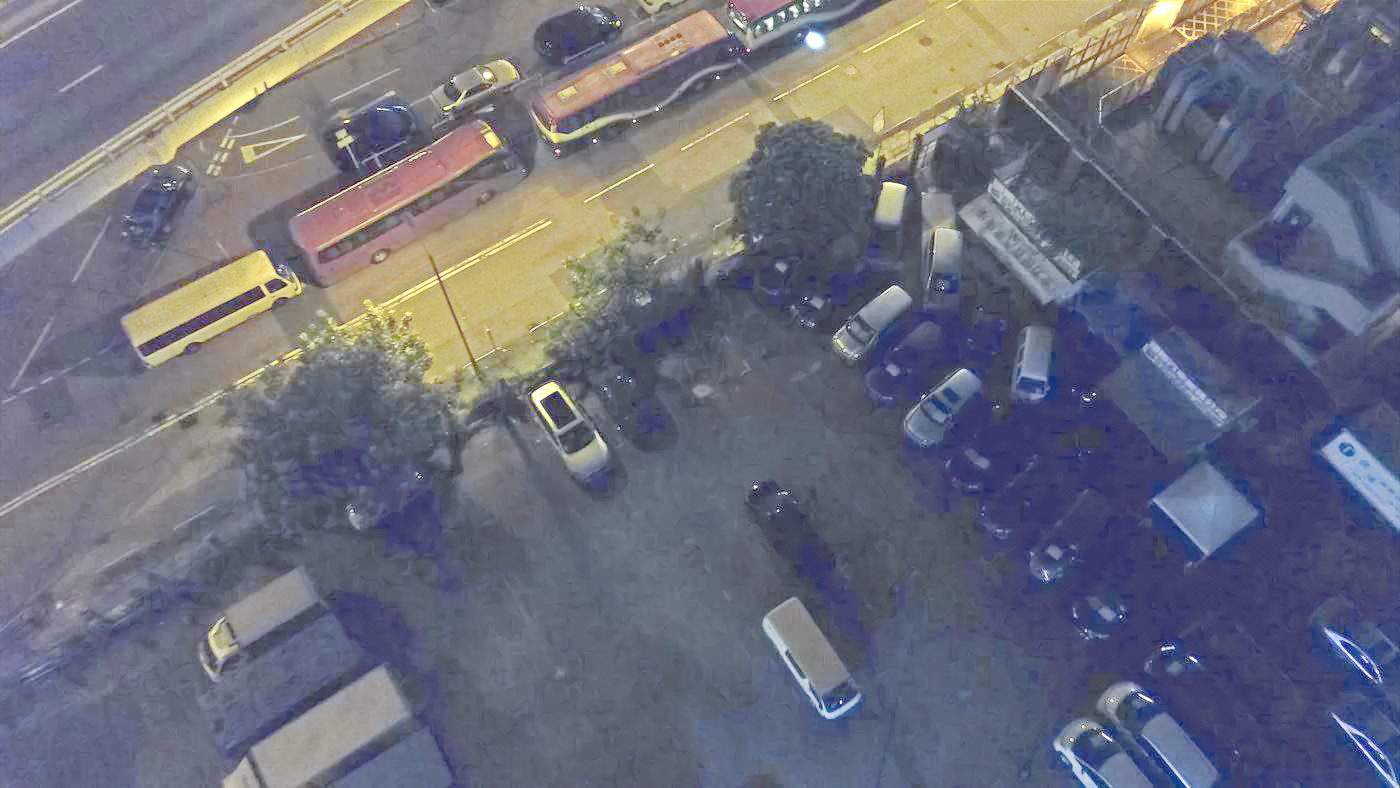}\vspace{-0.2em}
    \caption*{(d) URetinexNet~\cite{wu2022uretinexnet}}\medskip
    
    \includegraphics[width=1.\textwidth]{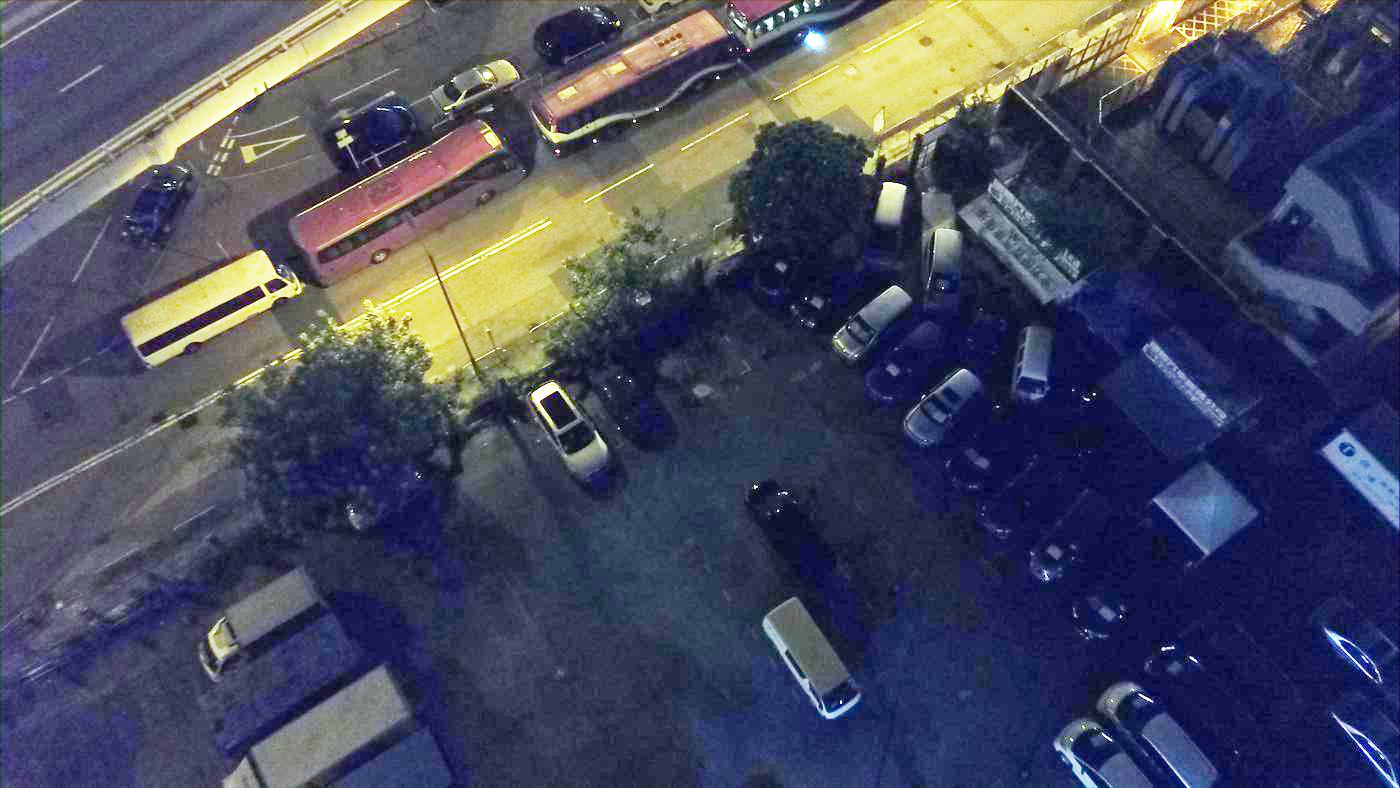}\vspace{-0.2em}
    \caption*{(j) SCI~\cite{ma2022sci}}\medskip
  \end{subfigure}
  \hfill
  \begin{subfigure}[c]{0.160\textwidth}
    \includegraphics[width=1.\textwidth]{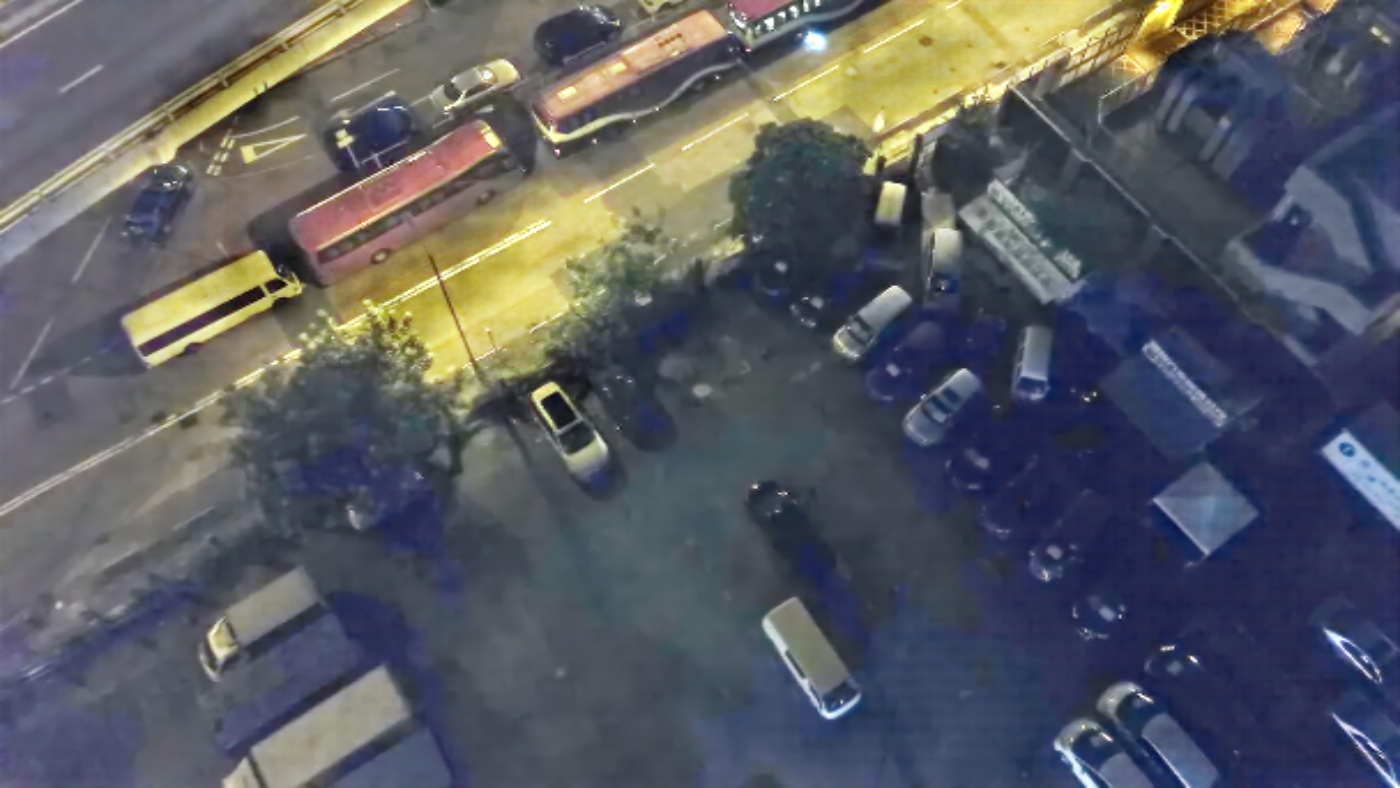}\vspace{-0.2em}
    \caption*{(e) SNR-Net~\cite{xu2022snr}}\medskip
    
    \includegraphics[width=1.\textwidth]{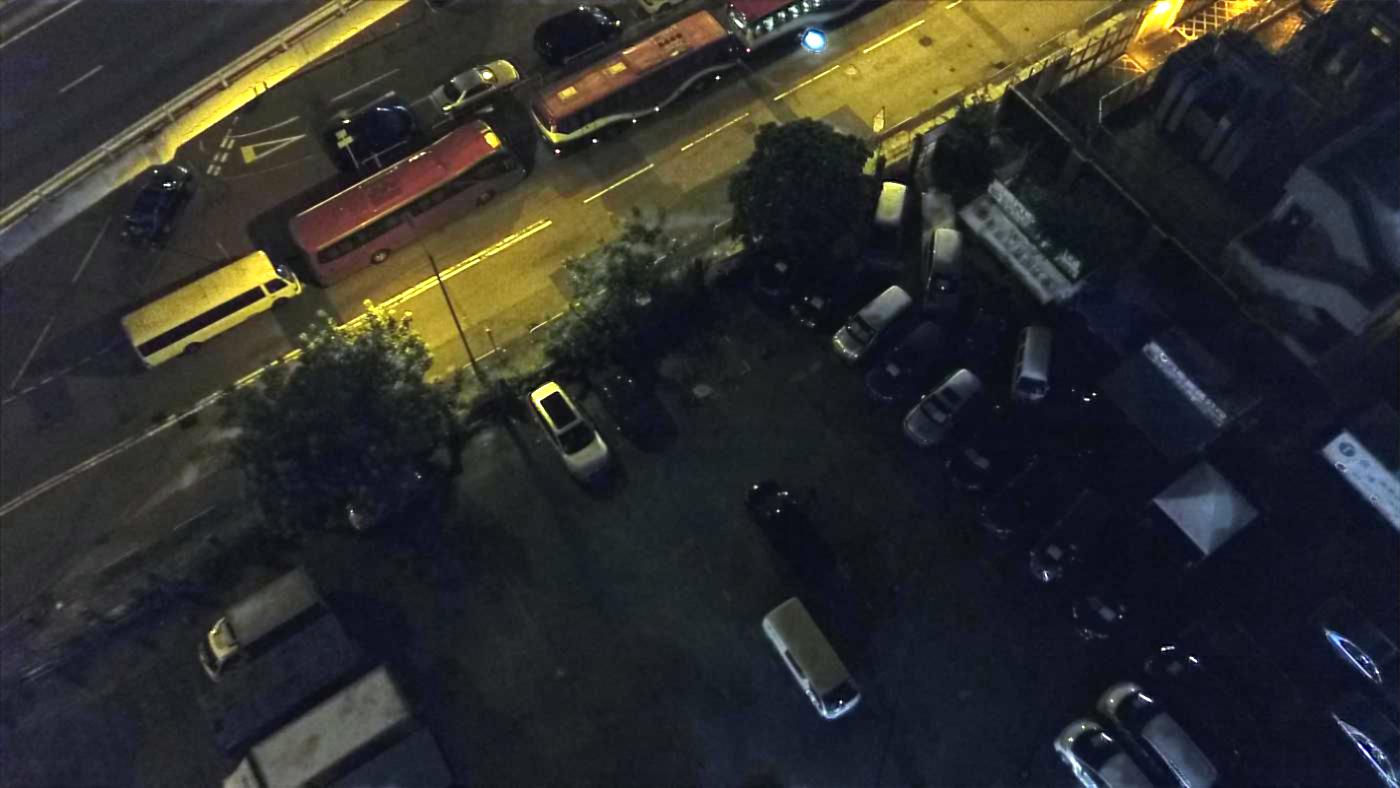}\vspace{-0.2em}
    \caption*{(k) NeRCo~\cite{yang2023implicit}}\medskip
  \end{subfigure}
  \begin{subfigure}[c]{0.160\textwidth}
    \includegraphics[width=1.\textwidth]{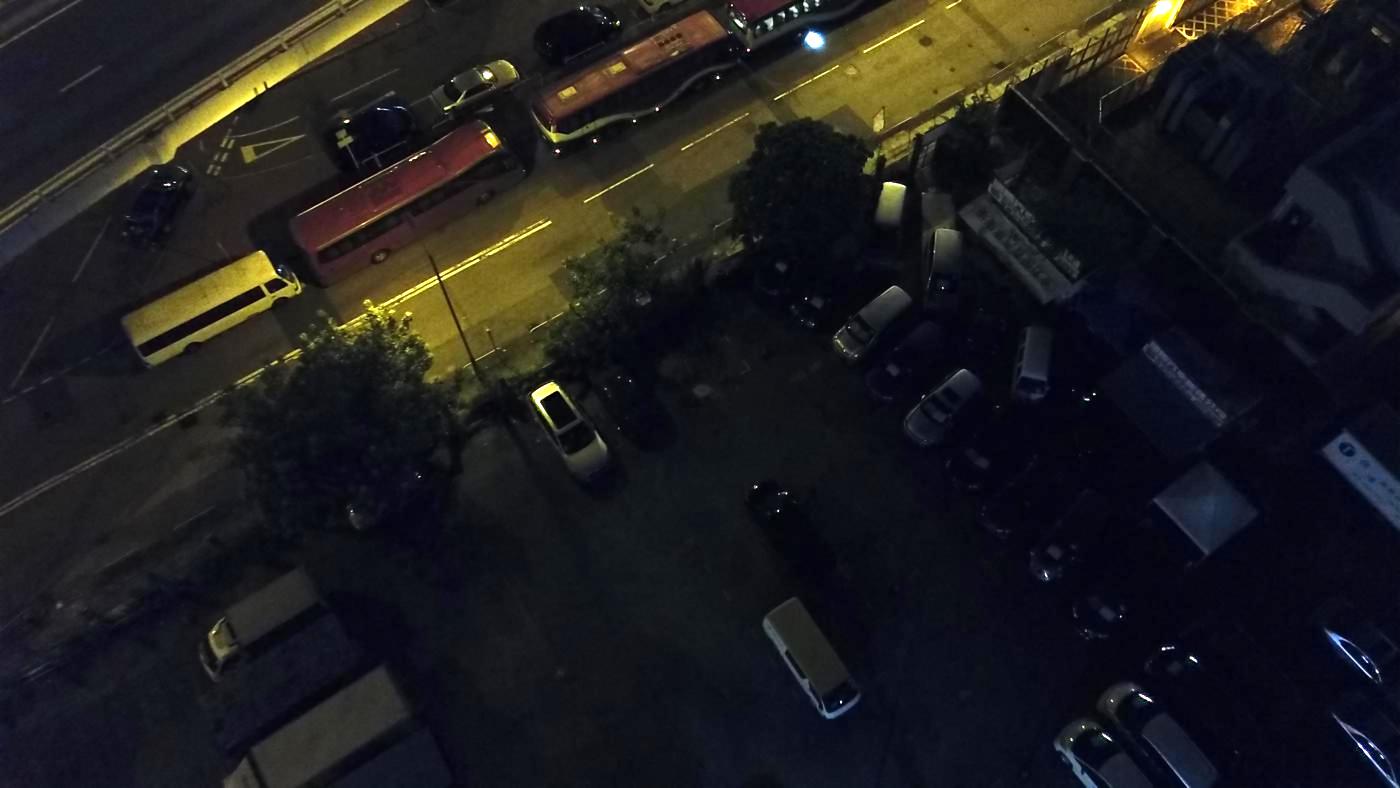}\vspace{-0.2em}
    \caption*{(f) Retinexformer~\cite{cai2023retinexformer}}\medskip
    
    \includegraphics[width=1.\textwidth]{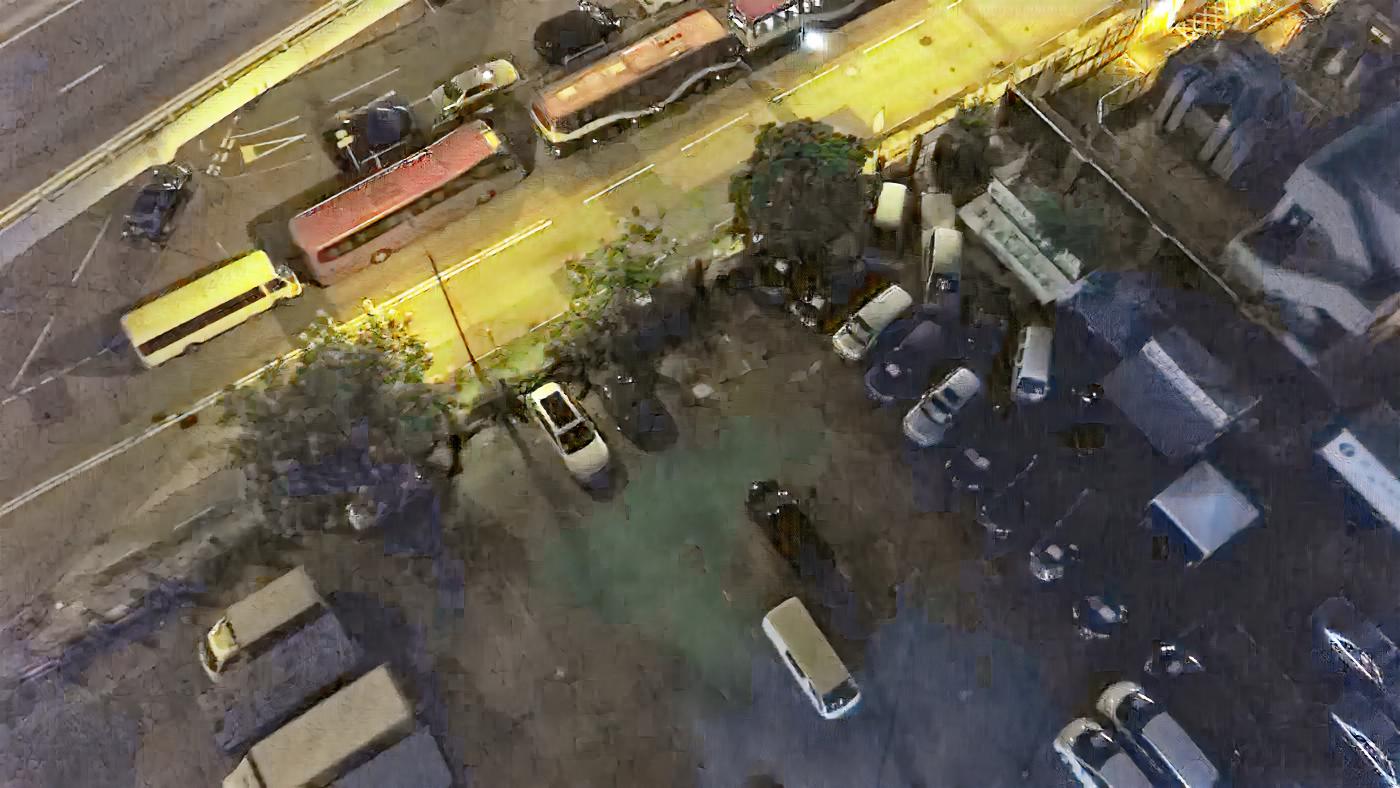}\vspace{-0.2em}
    \caption*{(l) Ours}\medskip
  \end{subfigure}
  \vspace{-0.8em}
  \caption{Visual comparison results of various methods on the unpaired test dataset of Visdrone. Best zoomed in for detail.}
  \label{fig:qualitative_comp_results}
    \vspace{-1.2em}
\end{figure*}

\begin{figure*}[t]
  \centering
  \begin{subfigure}[c]{0.160\linewidth}
    \includegraphics[width=1.\linewidth]{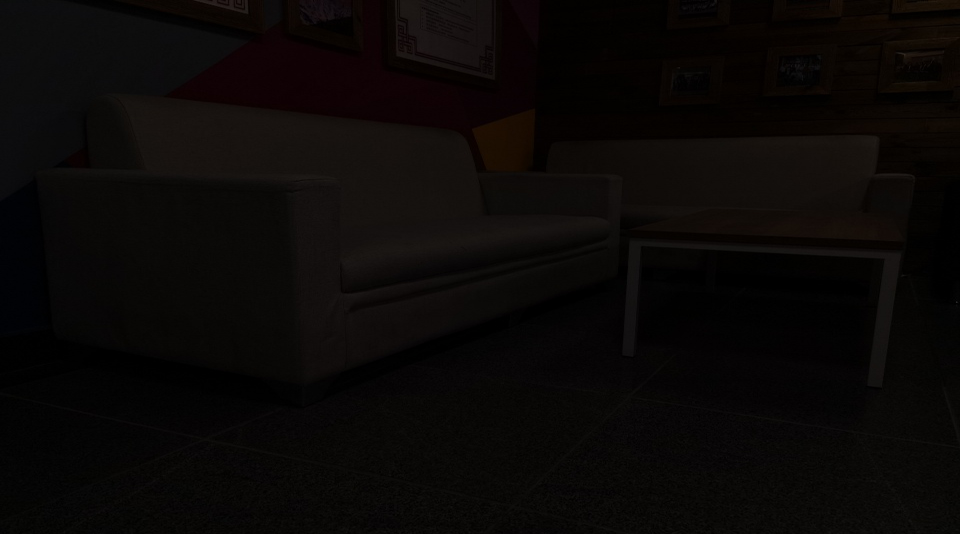}\vspace{-0.2em}
    \caption*{(a) Input}\medskip
    
    \includegraphics[width=1.\linewidth]{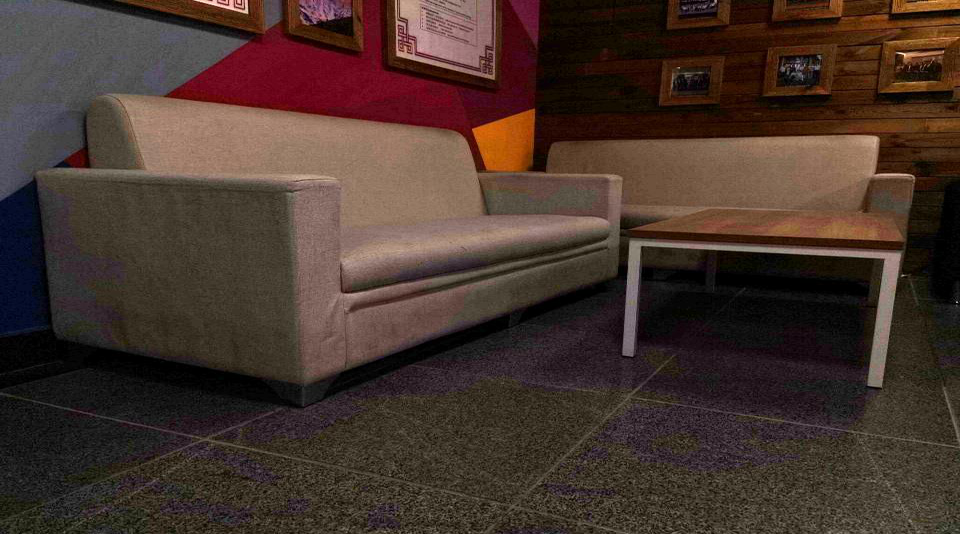}\vspace{-0.2em}
    \caption*{(g) RUAS~\cite{liu2021ruas} }\medskip
  \end{subfigure}
  \hfill
  \begin{subfigure}[c]{0.160\linewidth}
    \includegraphics[width=1.\linewidth]{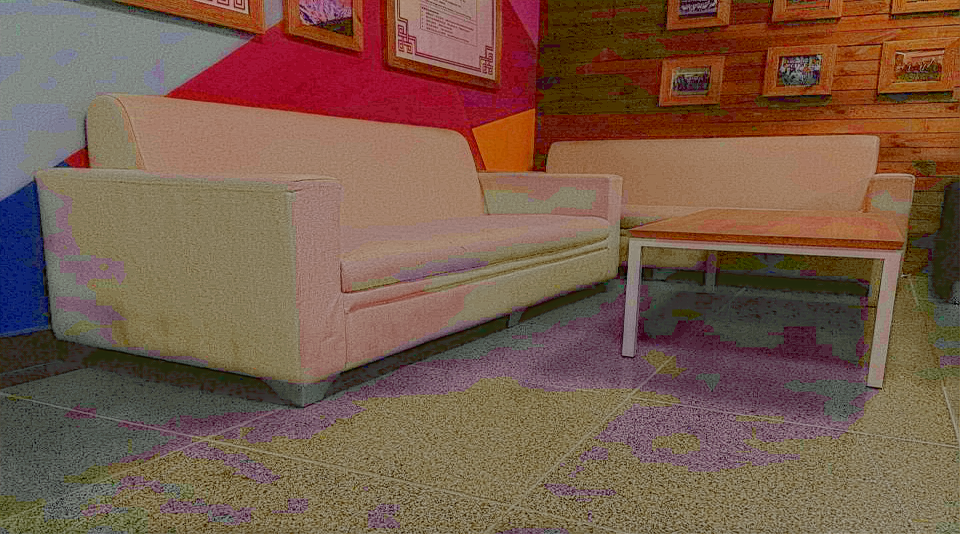}\vspace{-0.2em}
    \caption*{(b) RetinexNet~\cite{Chen2018Retinex}}\medskip
    
    \includegraphics[width=1.\linewidth]{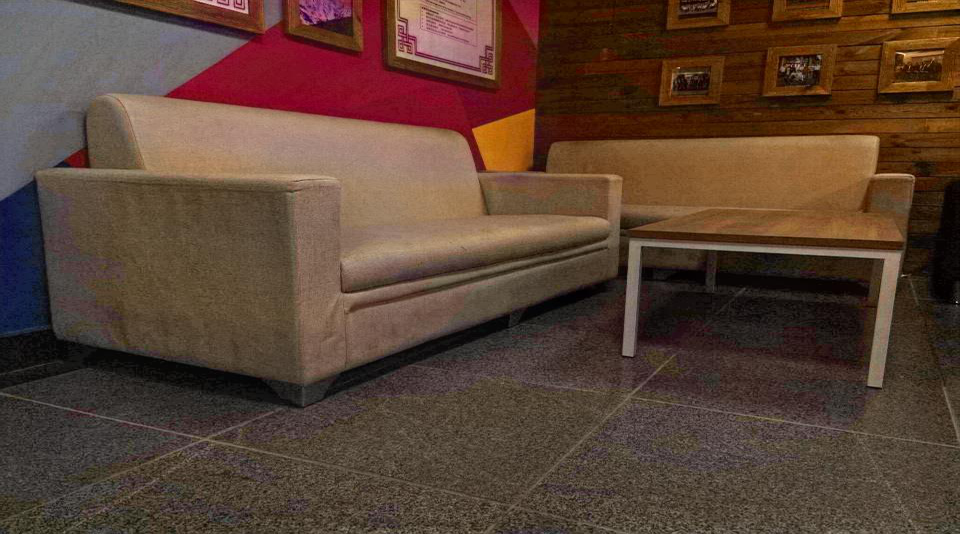}\vspace{-0.2em}
    \caption*{(h) EnGAN~\cite{jiang2021enlightengan}}\medskip
  \end{subfigure}
  \hfill
  \begin{subfigure}[c]{0.160\linewidth}
    \includegraphics[width=1.\linewidth]{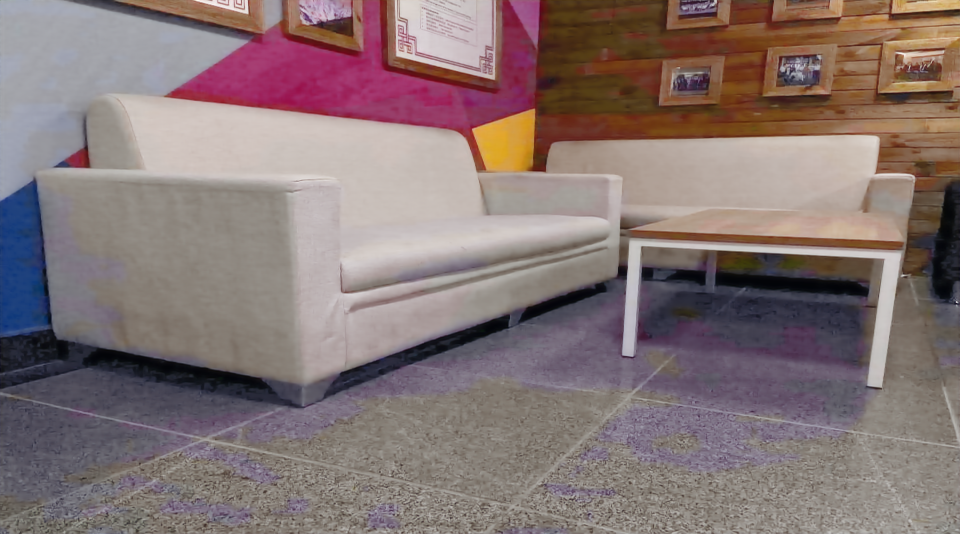}\vspace{-0.2em}
    \caption*{(c) MIRNet~\cite{zamir2020mirnet}}\medskip
    
    \includegraphics[width=1.\textwidth]{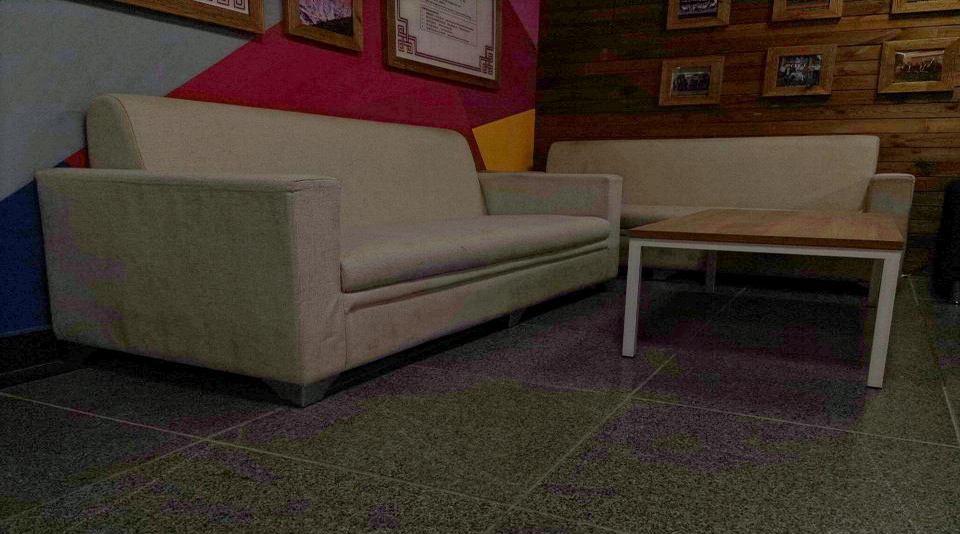}\vspace{-0.2em}
    \caption*{(i) ZeroDCE~\cite{guo2020zerodce}}\medskip
  \end{subfigure}
  \hfill
  \begin{subfigure}[c]{0.160\textwidth}
    \includegraphics[width=1.\textwidth]{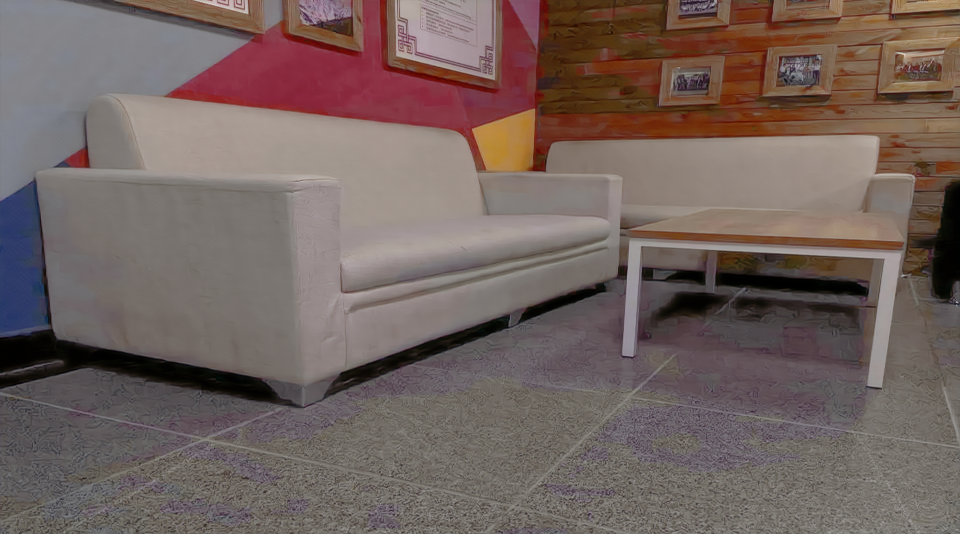}\vspace{-0.2em}
    \caption*{(d) URetinexNet~\cite{wu2022uretinexnet}}\medskip
    
    \includegraphics[width=1.\textwidth]{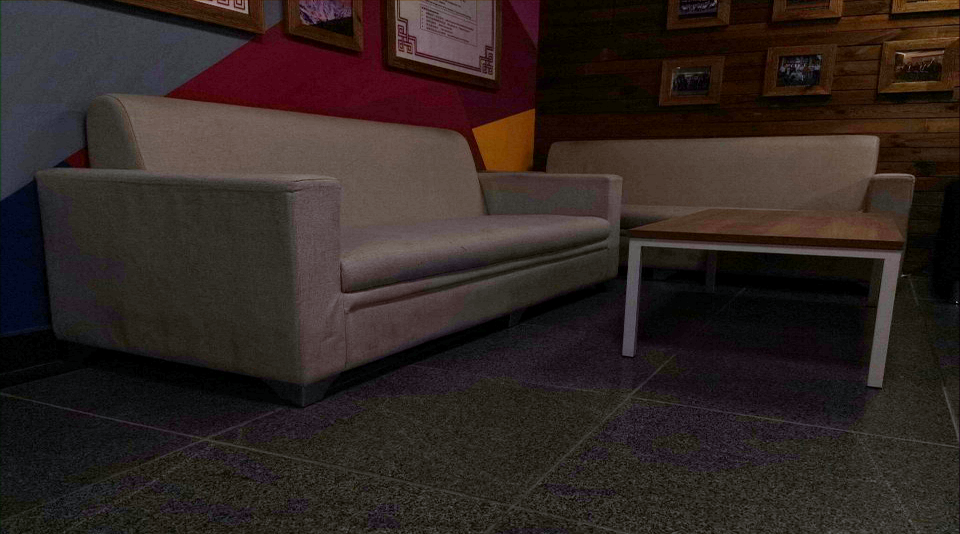}\vspace{-0.2em}
    \caption*{(j) SCI~\cite{ma2022sci}}\medskip
  \end{subfigure}
  \hfill
  \begin{subfigure}[c]{0.160\textwidth}
    \includegraphics[width=1.\textwidth]{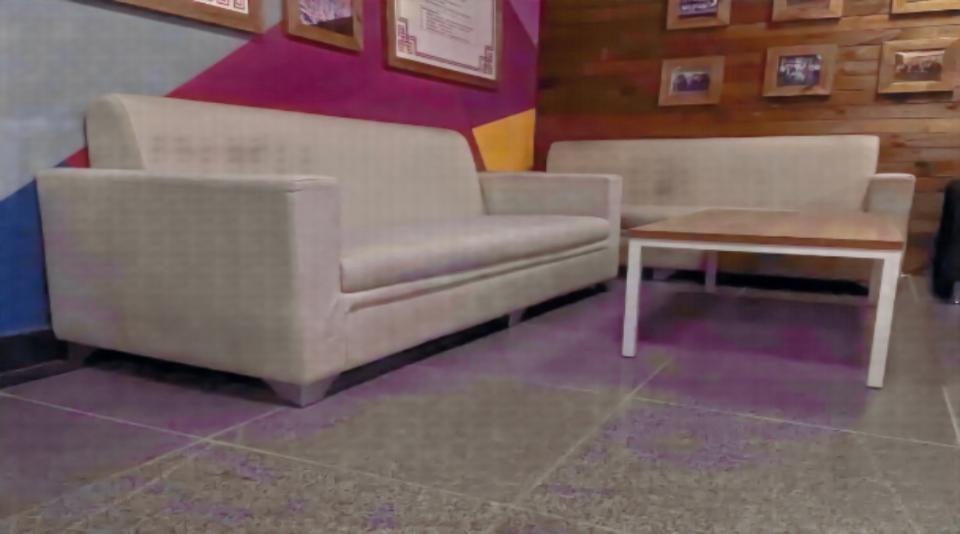}\vspace{-0.2em}
    \caption*{(e) SNR-Net~\cite{xu2022snr}}\medskip
    
    \includegraphics[width=1.\textwidth]{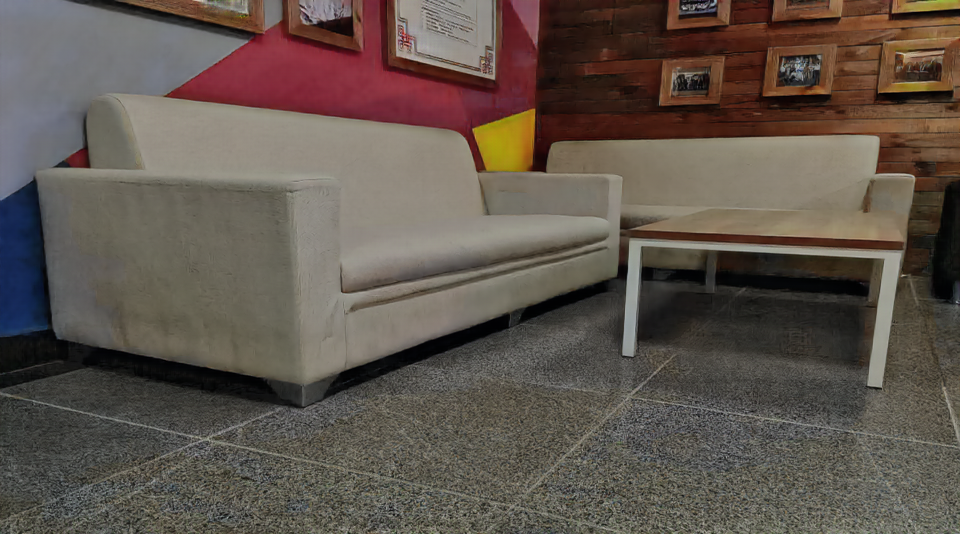}\vspace{-0.2em}
    \caption*{(k) Ours}\medskip
  \end{subfigure}
  \begin{subfigure}[c]{0.160\textwidth}
    \includegraphics[width=1.\textwidth]{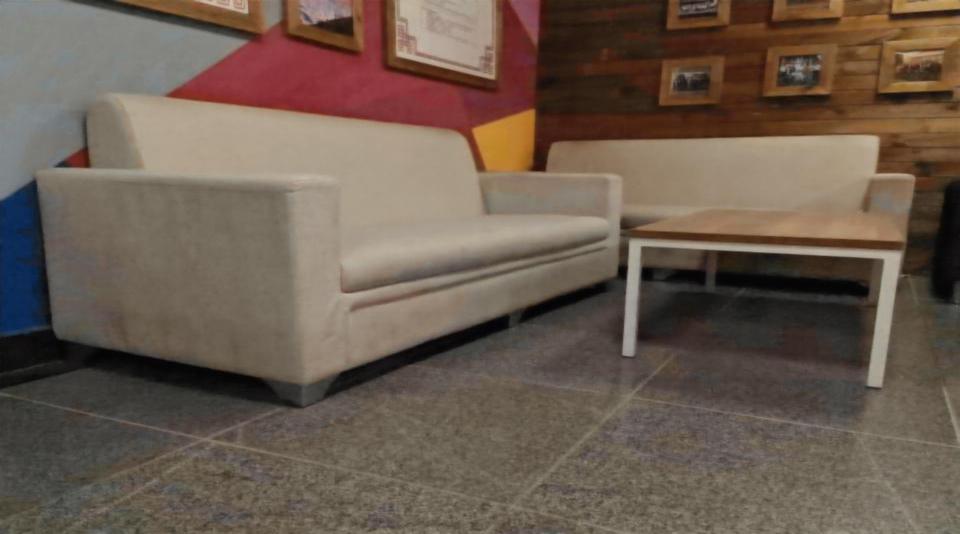}\vspace{-0.2em}
    \caption*{(f) Retinexformer~\cite{cai2023retinexformer}}\medskip
    
    \includegraphics[width=1.\textwidth]{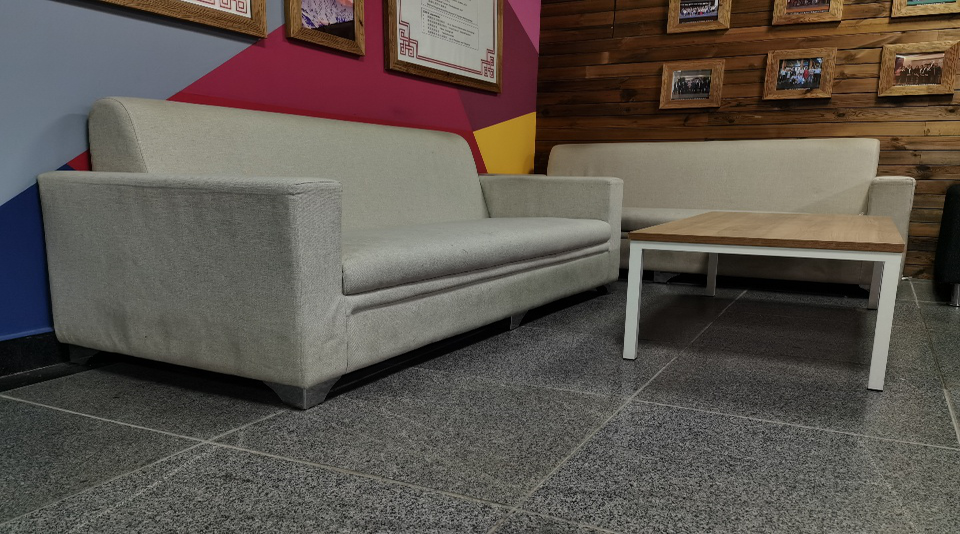}\vspace{-0.2em}
    \caption*{(l) GT}\medskip
  \end{subfigure}
  \vspace{-0.8em}

  \caption{Visual comparison results of various methods on the paired test dataset of LSRW. Best zoomed in for detail.}
  \label{fig:qualitative_comp_results_on_LSRW}
  \vspace{-1.2em}
\end{figure*}

\subsubsection{Qualitative Analysis}

We illustrate the visual comparison results of our Semi-LLIE and other SOTA methods on the Visdrone dataset in Fig.~\ref{fig:qualitative_comp_results}. The enhanced images generated by RetinexNet~\cite{Chen2018Retinex} suffer from heavy color distortion and severe textural artifacts. 
MIRNet~\cite{zamir2020mirnet}, RetinexFormer~\cite{cai2023retinexformer}, RUAS~\cite{liu2021ruas}, EnlightenGAN~\cite{jiang2021enlightengan} and NeRCo~\cite{yang2023implicit} fails to enhance such a hard scenario effectively and generate images with a darker tone. Notably, the images generated by the current SOTA-supervised Retinexformer~\cite{cai2023retinexformer} look even darker, which suggests that Retinexformer~\cite{cai2023retinexformer} can not generalize well to this kind of real scenario. UReinexNet~\cite{wu2022uretinexnet}, SNR-Net~\cite{xu2022snr}, ZeroDCE~\cite{guo2020zerodce}, and SCI~\cite{ma2022sci} can effectively promote the brightness of the input images. However, images generated by these methods exhibit varying degrees of color distortion, where the images tend to have an overall bluish tint. Conversely, our Semi-LLIE effectively enhances poor visibility and low contrast and generates visual-friendly images with natural color and detailed textures. Guided by semantic-aware contrastive loss and RAM-based perceptual losses, our Semi-LLIE generalizes well on such real-world scenes.

Fig.~\ref{fig:qualitative_comp_results_on_LSRW} demonstrates the visual comparison results derived from the LSRW dataset~\cite{hai2023r2rnet}, which contains extremely low-light images, thus posing significant challenges for enhancement. Most methods generate unpleasant outputs for extremely low-light input images.
In contrast to the ground truth (GT), RUAS~\cite{liu2021ruas}, ZeroDCE~\cite{guo2020zerodce}, and SCI~\cite{ma2022sci}, produce images with low brightness and insufficient enhancement. RetinexNet~\cite{Chen2018Retinex}, MIRNet~\cite{zamir2020mirnet}, URetinexNet~\cite{wu2022uretinexnet}, SNR-Net~\cite{xu2022snr}, and EnlightenGAN~\cite{jiang2021enlightengan} introduce significant color deviations and artifacts, adversely impacting the enhanced images' visual quality. Retinexformer~\cite{cai2023retinexformer} generates images with better brightness and color, however, it is over-smoothed which results in detained information lost. This is mainly caused by only using traditional $L_1$ for model optimization.
Our proposed Semi-LLIE exhibits superior visual performance in global brightness, color restoration, and detail preservation, closely resembling the GT.
\subsection{Ablation Study}
To systematically assess different aspects of our Semi-LLIE, perform a series of ablation studies on the Visdrone~\cite{zhu2021detection} and the LSRW~\cite{hai2023r2rnet} dataset.

\subsubsection{Evaluation on Major Design Components}
\begin{table}[ht]
	\centering
	\caption{Quantitative evaluation results generated by employing different components. IE refers to the Illumination estimation module of the Mamba-based low-light image enhancement backbone, and MS denotes the MSSB of the Mamba-based low-light image enhancement backbone. SCL refers to sematic-aware contrastive loss, and RP refers to RAM-based perceptual loss. The top result and the second-best result are marked in \textbf{bold} and \underline{underlined}, respectively.}
	\resizebox{\linewidth}{!}{
	\begin{tabular}{c|c c c c|ccc|cc}
		\toprule[1.5pt]
            \multirow{2}{*}{Method} & \multirow{2}{*}{IE} & \multirow{2}{*}{MS} & \multirow{2}{*}{SCL} & \multirow{2}{*}{RP}  & \multicolumn{3}{c|}{\cellcolor{gray!40}\emph{Visdrone Dataset}} & \multicolumn{2}{c}{\cellcolor{gray!40}\emph{LSRW Dataset}} \\ \cline{6-10}
            
              & & & &  & \cellcolor{gray!40}FID & \cellcolor{gray!40}NIQE $\downarrow$ & \cellcolor{gray!40}LOE $\downarrow$ &\cellcolor{gray!40}PSNR $\uparrow$ & \cellcolor{gray!40}SSIM $\uparrow$ \\ 
  
        \midrule[1.5pt]
		\#1 Base& \XSolidBrush & \XSolidBrush & \XSolidBrush & \XSolidBrush& 104.99 & 4.806 & 324.5 & 17.56 & 0.7166\\ \cline{6-10}
		\#2 Base+IE  & \CheckmarkBold & \XSolidBrush & \XSolidBrush & \XSolidBrush& 89.80 & 4.629 & 276.3 & 18.66 & 0.7625\\ \cline{6-10}
  		\#3 Base+MS  & \XSolidBrush & \CheckmarkBold & \XSolidBrush & \XSolidBrush & 83.00 & 4.009 & 268.5 & 18.70 & 0.7753\\ \cline{6-10}
            \#4 Base+IE+MS& \CheckmarkBold & \CheckmarkBold & \XSolidBrush & \XSolidBrush & 64.55 & 3.854 & 247.9 & 18.79 & 0.7794\\ \cline{6-10}
  		\#5 Base+IE+MS+SCL & \CheckmarkBold & \CheckmarkBold & \CheckmarkBold & \XSolidBrush & 52.69 & 3.825 & 231.9 & 19.22 & 0.7818\\ \cline{6-10}
  		\#6 Base+IE+MS+RP & \CheckmarkBold & \CheckmarkBold & \XSolidBrush & \CheckmarkBold & \underline{45.93} & \underline{3.767} & \underline{228.7} & \underline{19.39} & \underline{0.7836}\\ \cline{6-10}
		\#7 Semi-LLIE (Ours)& \CheckmarkBold & \CheckmarkBold & \CheckmarkBold & \CheckmarkBold & \textbf{36.49} & \textbf{3.667} & \textbf{204.1} & \textbf{19.73} & \textbf{0.7840} \\ 
        \bottomrule[1.5pt]
	\end{tabular}
	}

        \label{tab:ablation_major_comp}
	\vspace{-1.em}
\end{table}





To investigate the individual contributions of the primary components of our Semi-LLIE, namely the illumination estimation module, the multi-scale state space block, semantic-aware contrastive loss, and the RAM-based perceptual loss, we explore several configurations by progressively integrating the proposed components into the baseline architecture. They are listed as follows:
\begin{enumerate}

\item  "\#1 Base" refers to the basic semi-supervised learning strategy with $L_1$ consistency loss and introduces the MambaIR~\cite{guo2024mambair} as our enhancement backbone.

\item  "\#2 Base+IE" integrates an illumination estimation module into the MambaIR~\cite{guo2024mambair} enhancement backbone, aiming to generate a light-up illumination map to guide the enhancement process.

\item  "\#3 Base+MS" replaces the VSSB in MambaIR~\cite{guo2024mambair} enhancement backbone with our multi-scale state space block (MSSB), aiming to compare the performance gains of MSSB over VSSB in MabaIR~\cite{guo2024mambair}. 

\item  "\#4 Base+IE+MS" integrates both the illumination estimation module and our multi-scale state space block (MSSB) into MambaIR~\cite{guo2024mambair} enhancement backbone. 

\item  "\#5 Base+IE+MS+SCL" refers to training the "Baseline+IE+MM" by introducing the additional semantic-aware contrastive loss, leveraging the strong image representation learning ability of RAM's image Encoder. 

\item  "\#6 Base+IE+MS+RP" refers to training the "Baseline+IE+MM" by introducing the additional RAM-based perceptual loss, aiming to facilitate the generation of enhanced images with realistic details.

\item  Ultimately, we adopt a comprehensive "\#7 Semi-LLIE" approach by integrating all the primary components.
\end{enumerate}

We summarize the quantitative results of different configurations in Table~\ref{tab:ablation_major_comp}. 
As shown, our full configuration "\#7 Semi-LLIE" demonstrates a significant performance improvement against the baseline model "\#1 Base", which benefits from each of the major components of Semi-LLIE. Specifically, comparisons of "\#2 Base+IE" and "\#3 Base+MS" with "\#1 Base" demonstrate that integrating the illumination estimation module and multi-scale state space modules significantly enhances overall model performance. Further comparison of "\#4 Base+IE+MS" with "\#1 Base" indicates that combining both modules yields additional performance improvements in the enhancement backbone. The "\#4 Base+IE+MS" refers to our Mamba-based low-light image enhancement backbone, which outperforms the baseline model "\#1 Base" in all three non-referenced metrics on the Visdrone dataset~\cite{zhu2021detection}. It suggests that our Mamba-based low-light image enhancement backbone produces visual-friendly enhanced images. On the LSRW datasets, our Mamba-based low-light image enhancement backbone achieves a $1.23$ dB gain in PSNR and $0.0628$ in SSIM, which suggests that our Mamba-based low-light image enhancement backbone enables us to restore more details buried in the under-exposed images.
By comparing "\#5 Baseline+IE+MM+SCL" with "\#4 Baseline+IE+MM", it is apparent to validate the effectiveness of incorporating the semantic-aware contrastive loss, where "\#5" outperforms "\#4" in all three metrics on the Visdrone dataset~\cite{zhu2021detection} while obtaining $0.43$ dB gains in PSNR and $0.024$ in SSIM on the LSRW Dataset. 
By comparing "\#6 Baseline+IE+MM+RP" with "\#4 Baseline+IE+MM", the significance of RAM-based perceptual loss can be confirmed.

\begin{figure}[t]
  \centering
  \begin{subfigure}{0.24\linewidth}
    \includegraphics[width=1.\linewidth]{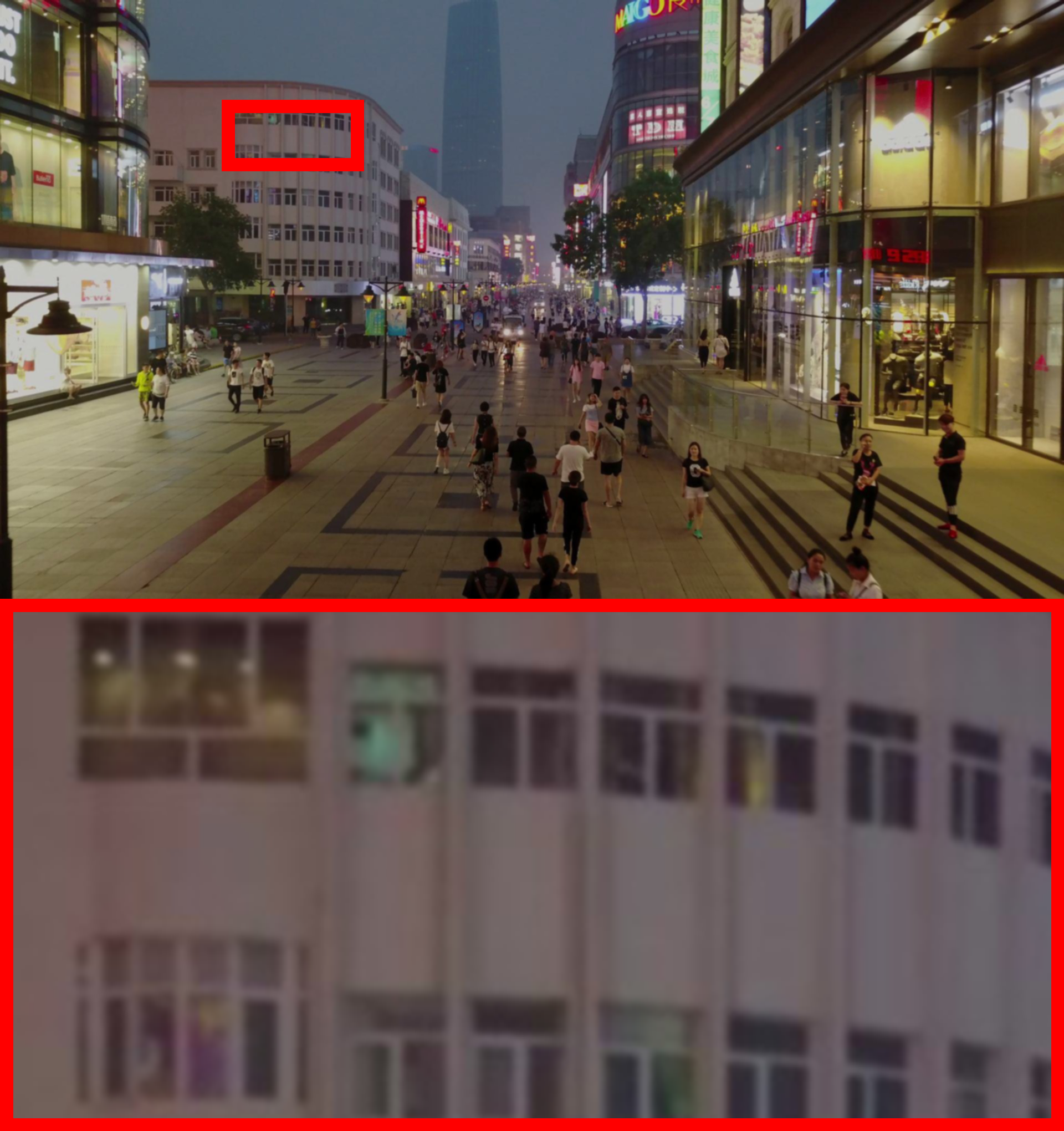}
    \caption*{(a) Input}\medskip
    
    \includegraphics[width=1.\linewidth]{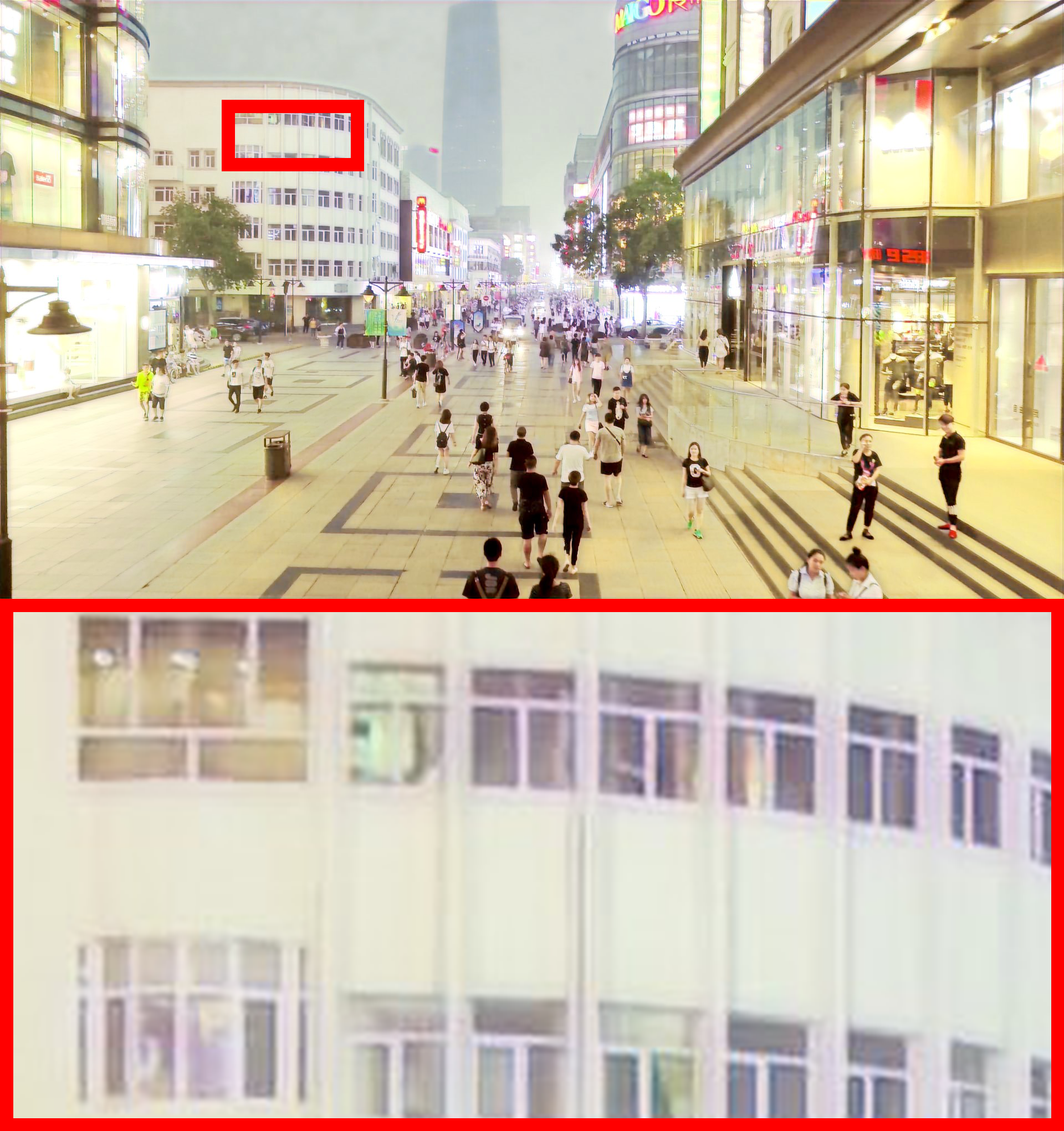}
    \caption*{ (d) \#4}
  \end{subfigure}
  \hfill
  \begin{subfigure}{0.24\linewidth}
    \includegraphics[width=1.\linewidth]{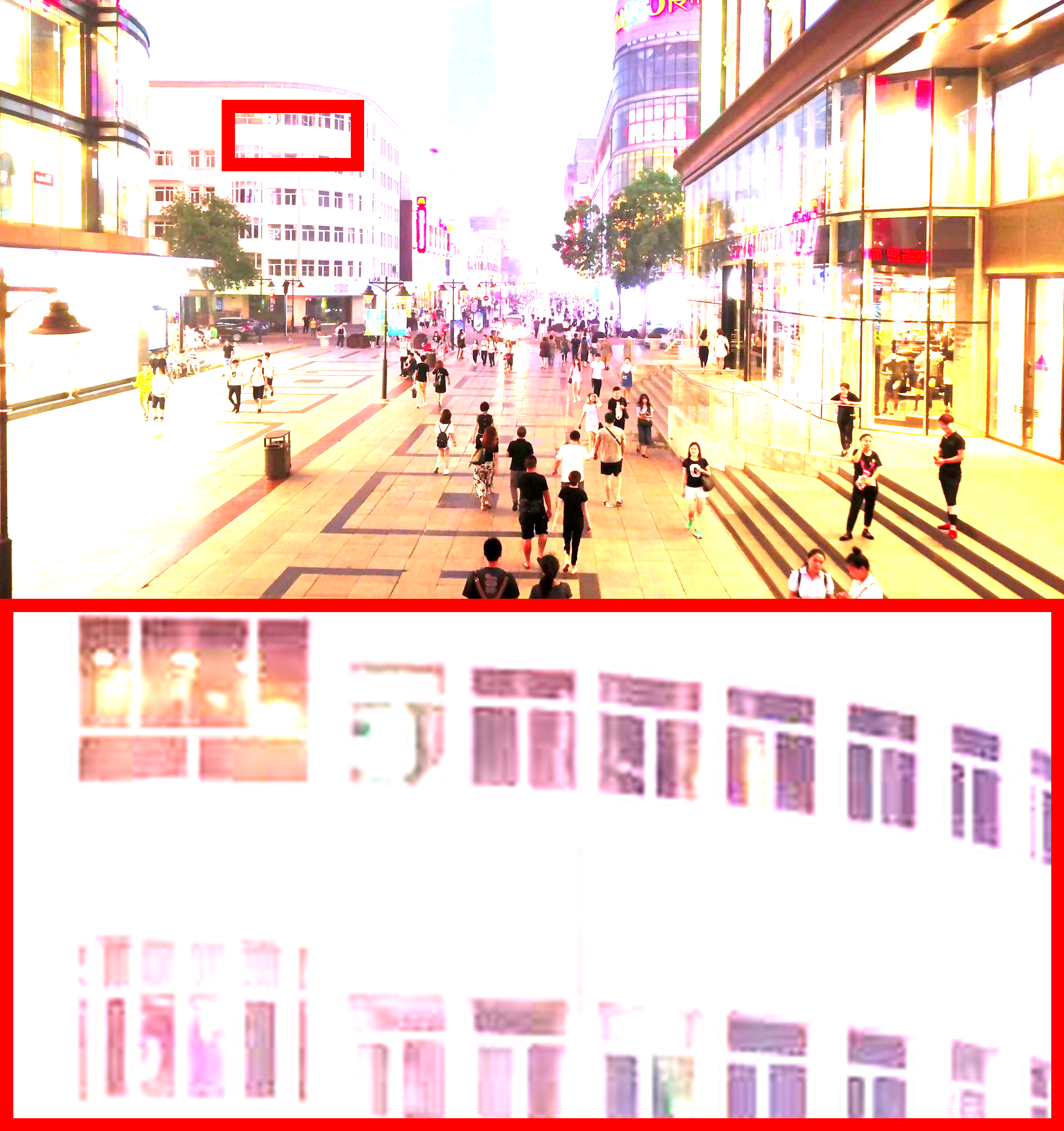}
    \caption*{(b) \#1 }\medskip
    
    \includegraphics[width=1.\linewidth]{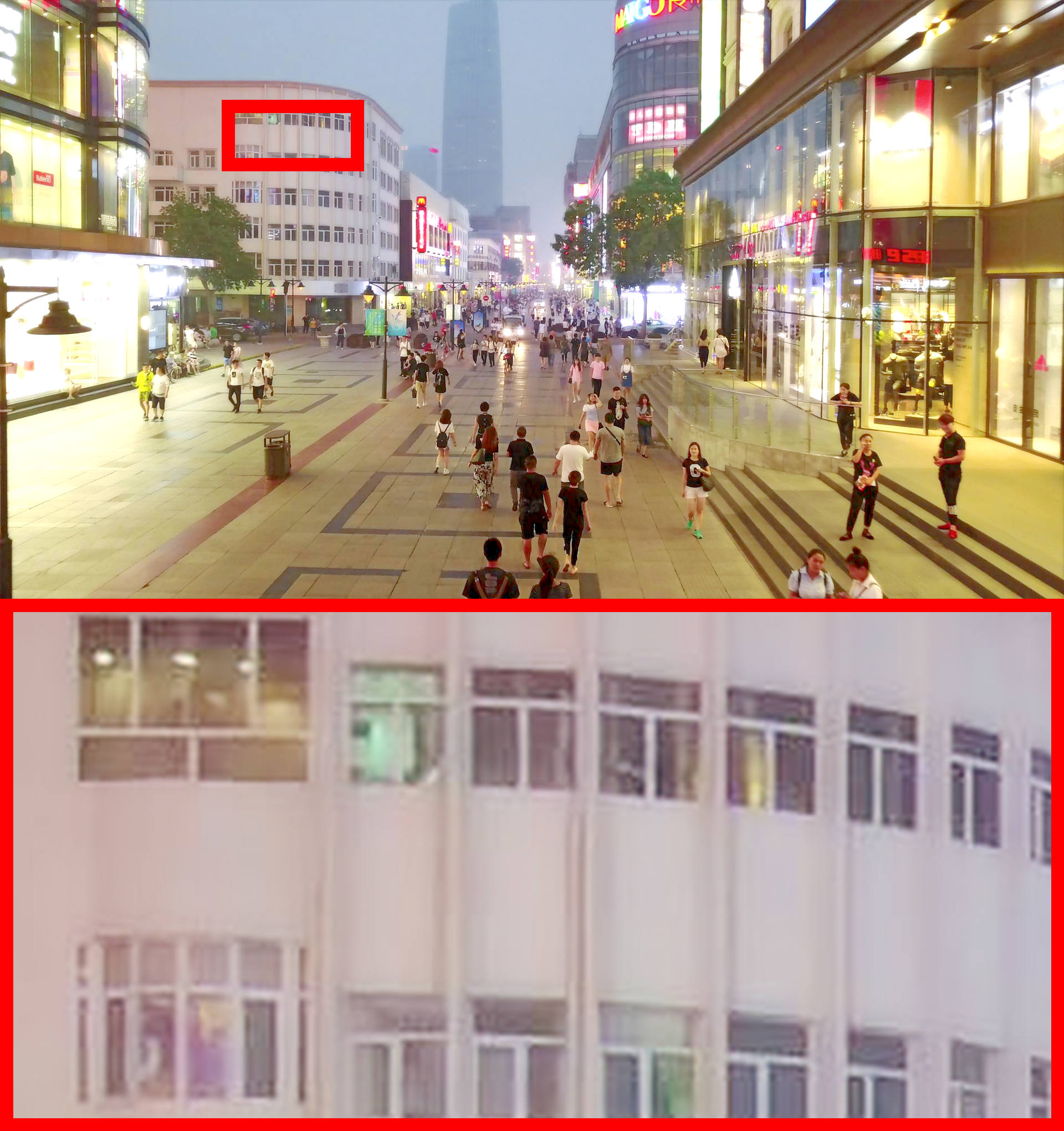}
    \caption*{(e) \#5}
  \end{subfigure}
  \hfill
  \begin{subfigure}{0.24\linewidth}
    \includegraphics[width=1.\linewidth]{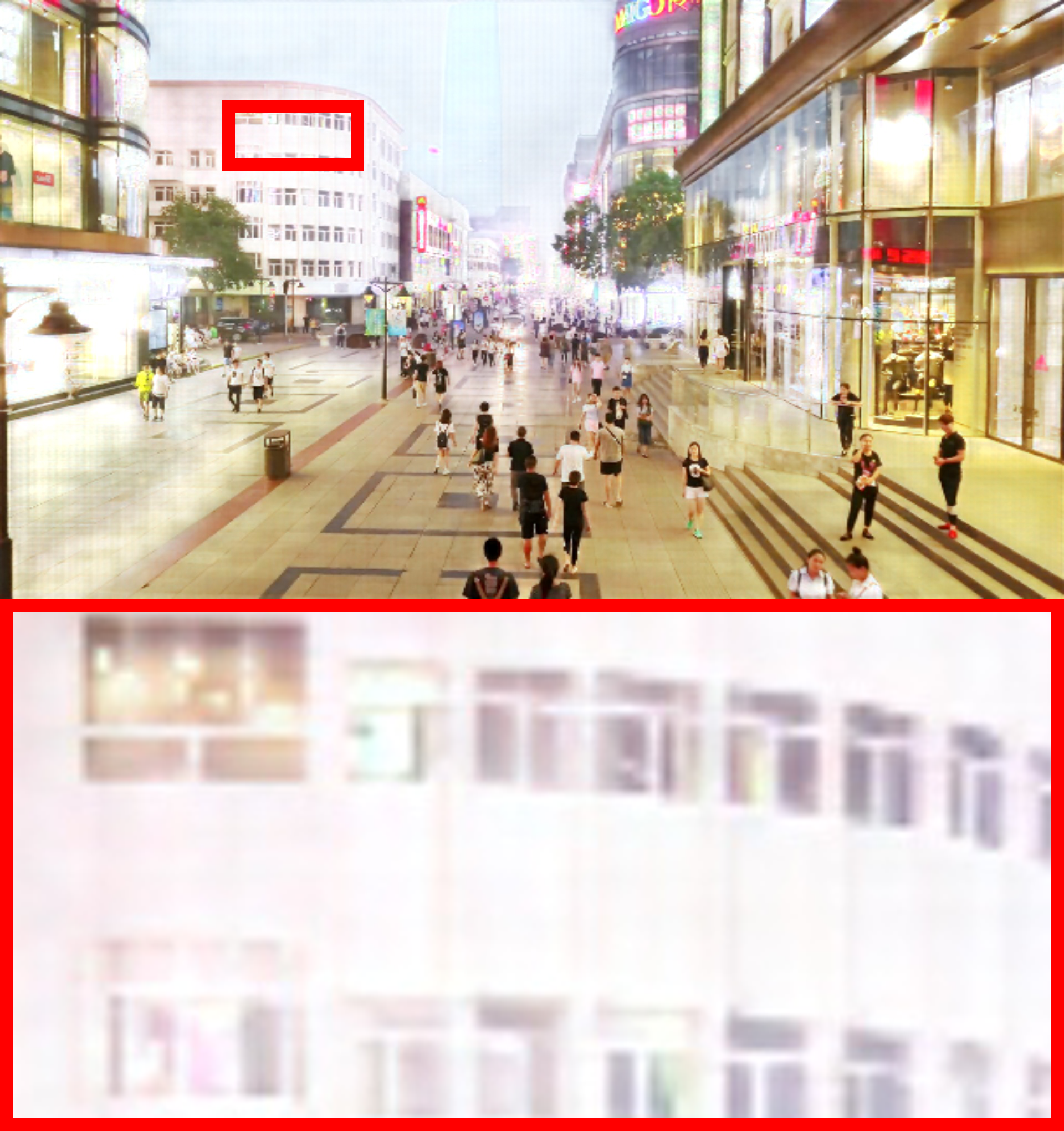}
    \caption*{(c) \#2 }\medskip
    
    \includegraphics[width=1.\linewidth]{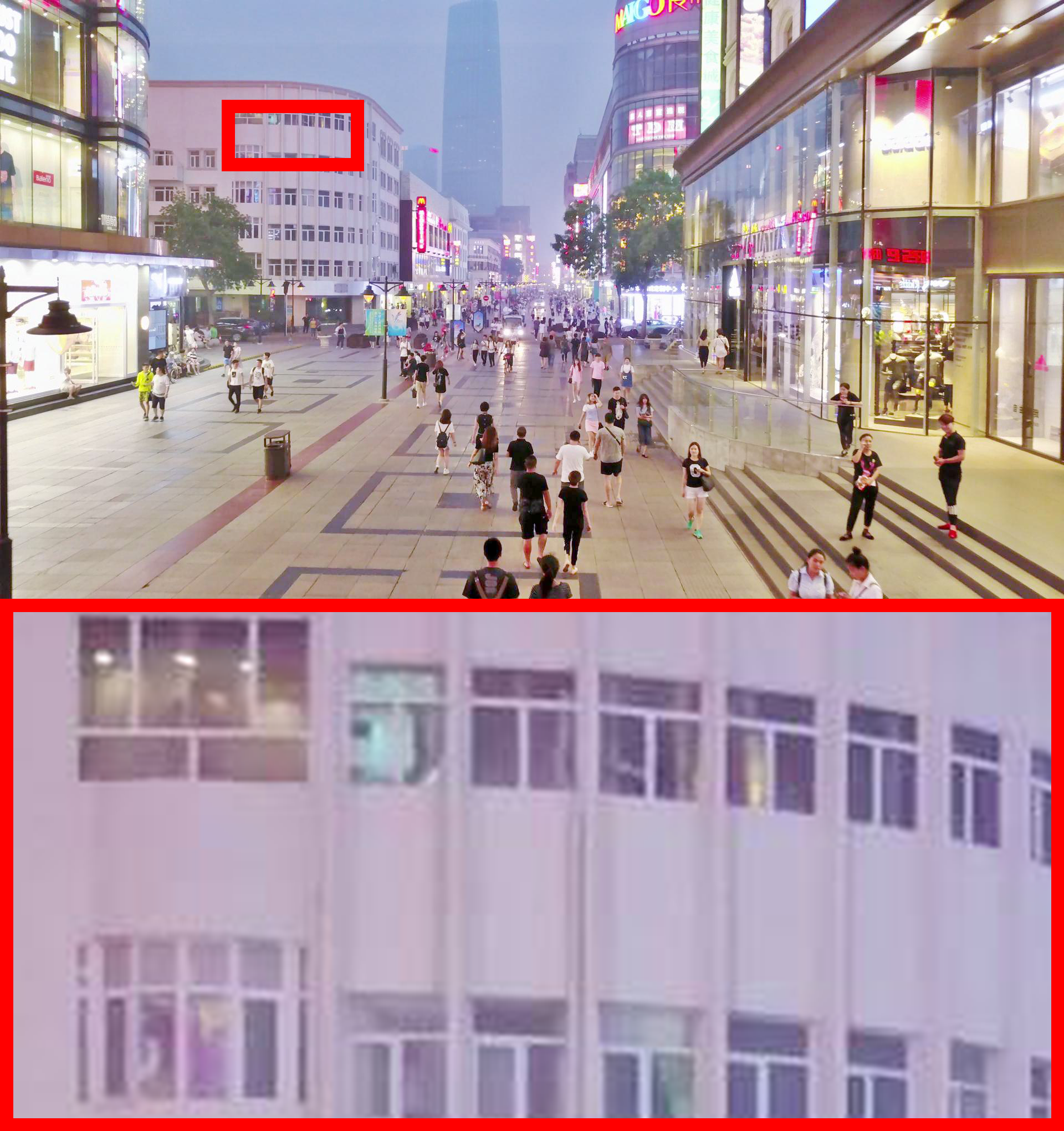}
    \caption*{(f) \#6}
  \end{subfigure}
  \hfill
  \begin{subfigure}{0.24\linewidth}
    \includegraphics[width=1.\linewidth]{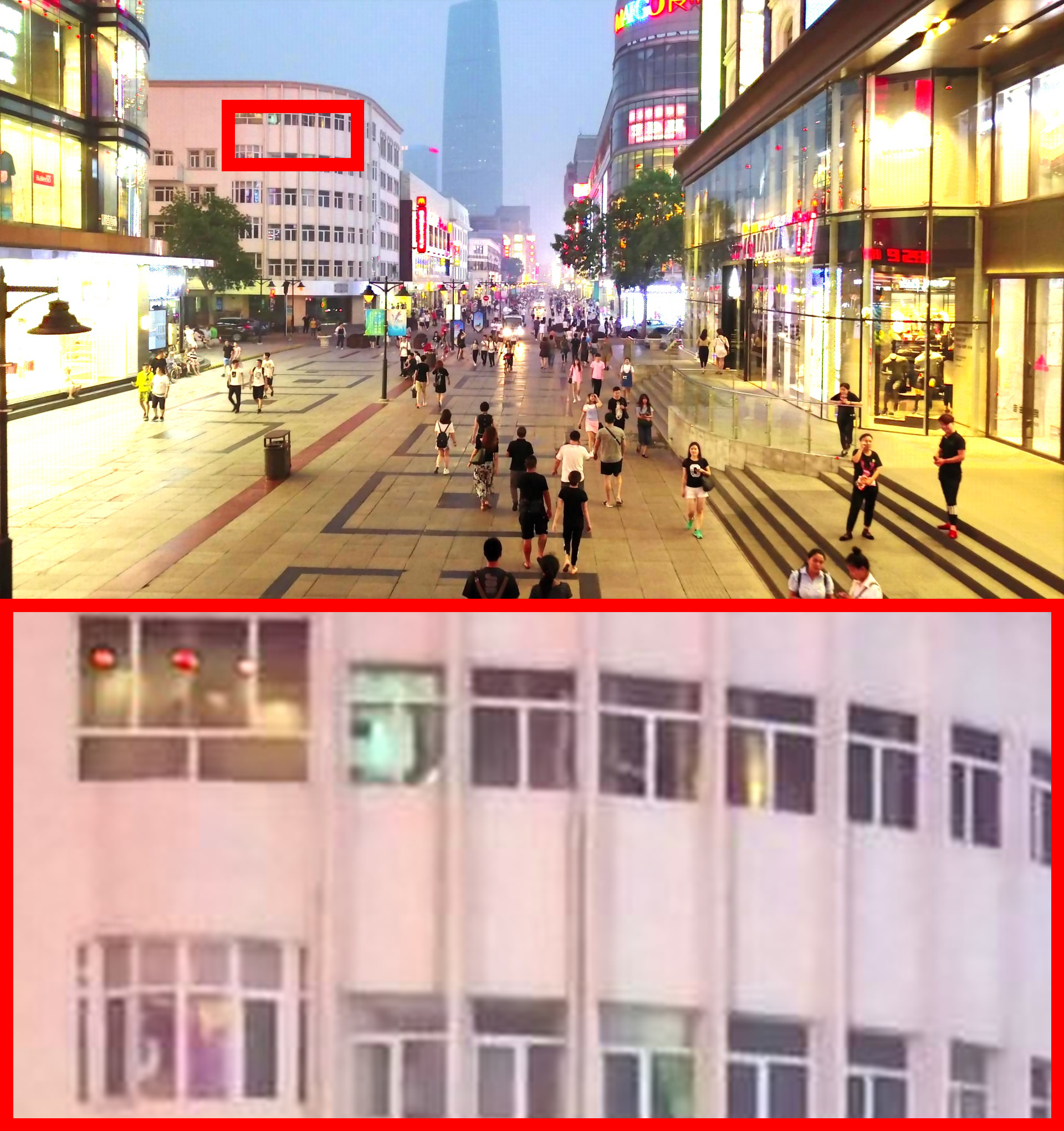}
    \caption*{(d) \#3 }\medskip
    
    \includegraphics[width=1.\linewidth]{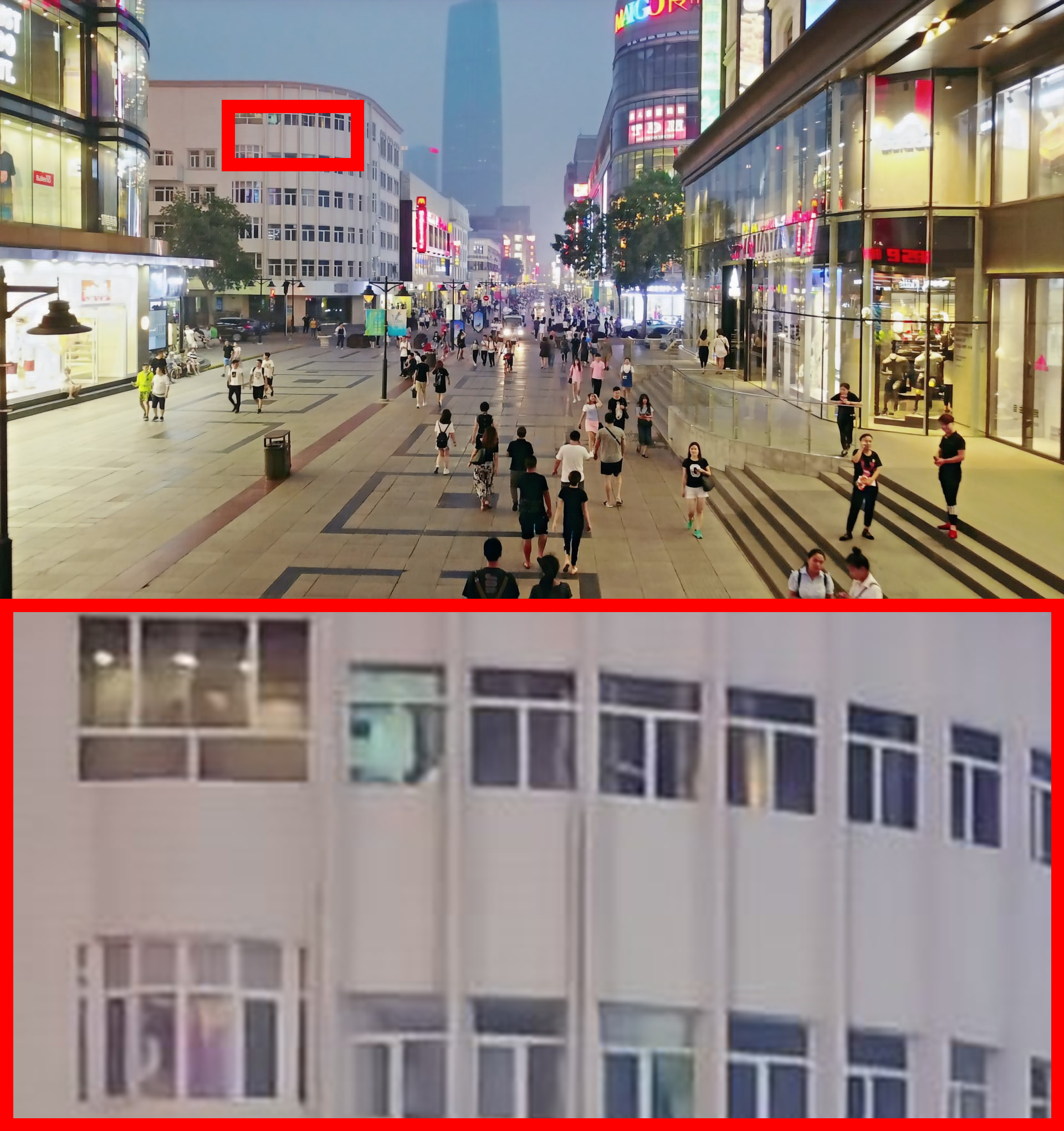}
    \caption*{(f) Semi-LLIE}
  \end{subfigure}
  \caption{Visualization of ablation results. The full configuration (f) performs best, particularly in the areas boxed in \textbf{\textcolor{red}{red}}.}
  \label{fig:qualitative_ablation}
  \vspace{-1.0em}

\end{figure}

We demonstrate the input low-light image and qualitative results of all ablation configurations in Fig.~\ref{fig:qualitative_ablation}.
As shown, models "\#1" and  "\#2" can enhance the contrast and brightness of the image to some degree, but significant over-exposure and blurriness problems exist. Owing to the multi-scale feature representation ability of MSSB, the enhanced images produced by "\#3" and  "\#4" tend to restore finer details from the input low-light image, and the issue of overexposure has been alleviated. However, the colors or tones are still inauthentic. By introducing the semantic-aware contrastive loss and the RAM-based perceptual loss, "\#5" and  "\#6" significantly mitigate the over-exposure issue and alleviate the color bias to a certain extent. With our full setting, the Semi-LLIE method effectively preserves image details and exhibits visual-friendly natural colors, especially in areas highlighted by red boxes. These outcomes can be attributed to the Mamba-based low-light image enhancement backbone, the semantic-aware contrastive loss, and the RAM-based perceptual loss. Our Mamba-based low-light image enhancement backbone extracts multi-scale features effectively by introducing MSSB. Based on the image encoder of the large-scale vision model RAM, the latter two newly designed losses perceive rich semantic information thus facilitating a perceptual-friendly optimization process.

\subsubsection{Evaluation on Image Encoders for Semantic-aware Contrastive Regularization}
\begin{table}[ht]
	\centering
	\caption{Evaluation on different pre-trained image encoders.} 
	\resizebox{0.9\linewidth}{!}{
	\begin{tabular}{c|ccc|cc}
		\toprule[1.pt]
            \multirow{2}{*}{Image Encoders}  & \multicolumn{3}{c|}{\cellcolor{gray!40}\emph{Visdrone Dataset}} & \multicolumn{2}{c}{\cellcolor{gray!40}\emph{LSRW Dataset}} \\ \cline{2-6}	
             
		& \cellcolor{gray!40}FID $\downarrow$ & \cellcolor{gray!40}NIQE $\downarrow$ & \cellcolor{gray!40}LOE $\downarrow$ & \cellcolor{gray!40}PSNR $\uparrow$ & \cellcolor{gray!40}SSIM $\uparrow$ \\ 
            
            \midrule
		VGG~\cite{simonyan2015very}  & 67.93 & 4.324 & 289.1 & 18.45  & 0.7752 \\ 
		  CLIP~\cite{radford2021learning}  & 48.65 & 3.826 & 247.0 & 18.89 & 0.7805 \\ 
		  SAM~\cite{kirillov2023segment}  & \underline{41.39} & \underline{3.823} & \underline{233.8} & \underline{19.04} & \underline{0.7828} \\ 
            
            \midrule

		RAM~\cite{zhang2024recognize} (Ours)  & \textbf{36.49} & \textbf{3.667} & \textbf{204.1}  & \textbf{19.73} & \textbf{0.7840} \\ 
        \bottomrule[1.pt]
	\end{tabular}}
	\vspace{-1.em}
	\label{tab:abl_image_encoders} 
\end{table}
We investigate the effectiveness of varied image encoders from pre-trained models for calculating the feature distance in contrastive regularization. We mainly analyze four pre-trained image encoders from VGG~\cite{simonyan2015very}, CLIP~\cite{radford2021learning}, SAM~\cite{kirillov2023segment}, and RAM~\cite{zhang2024recognize}. The comparison results are reported in Table~\ref{tab:abl_image_encoders}. Compared to the VGG-based image encoder, image encoders from models (CLIP~\cite{radford2021learning}, SAM~\cite{kirillov2023segment}, and RAM~\cite{zhang2024recognize}) trained with large-scale datasets exhibit more powerful semantic learning ability, leading to better performances. Besides, our RAM-based image encoder achieves the best performance, due to the refined large-scale image-text training pairs and the Swin Transformer~\cite{liu2021swin} backbone. Thus, we design our semantic-aware contrastive loss based on the image encoder of RAM.

\subsubsection{Evaluation on Enhancement Backbones}
\begin{table}[ht]
	\centering
	\caption{Investigation on current SOTA image enhancement baselines. }	
        \resizebox{0.9\linewidth}{!}{
	\begin{tabular}{c|ccc|cc}
		\toprule[1.pt]
              \multirow{2}{*}{Enhancement Baselines}  & \multicolumn{3}{c|}{\cellcolor{gray!40}\emph{Visdrone Dataset}} & \multicolumn{2}{c}{\cellcolor{gray!40}\emph{LSRW Dataset}} \\ \cline{2-6}	
              
		& \cellcolor{gray!40}FID $\downarrow$ & \cellcolor{gray!40}NIQE $\downarrow$ & \cellcolor{gray!40}LOE $\downarrow$ & \cellcolor{gray!40}PSNR $\uparrow$ & \cellcolor{gray!40}SSIM $\uparrow$ \\ 
		\midrule[1.pt]

		Restormer~\cite{zamir2022restormer}  & 67.11 & 4.507 & 243.98 & 18.66  & 0.7625\\ 
	    Uformer~\cite{wang2022uformer}  & 59.87 & 4.178 & 240.55 & 18.88 & 0.7801\\ 
		Retinexformer~\cite{cai2023retinexformer}  & 54.36 & 3.945 & 229.21 & 19.07 & 0.7819 \\ 
  	    MambaIR~\cite{guo2024mambair}  & \underline{48.52} & \underline{3.865} & \underline{223.47} & \underline{19.19} & \underline{0.7832} \\ 
            \midrule
		Ours  & \textbf{36.49} & \textbf{3.667} & \textbf{204.1}  & \textbf{19.73} & \textbf{0.7840} \\ 
        \bottomrule[1.pt]
	\end{tabular}}

    \label{tab:abl_LLIE_SOTAs}  
    \vspace{-1.em}	
\end{table}

We have conducted experiments to assess the efficacy of recently developed outstanding enhancement backbones by replacing our Mamba-based low-light image enhancement backbone with them in our Semi-LLIE framework without modifying the training losses. We summarize the comparison results in Table~\ref{tab:abl_LLIE_SOTAs}, where we can conclude several observations. Firstly, our Mamba-based low-light image enhancement backbone significantly outperforms the Transformer-based methods without an illumination estimation module (Restormer~\cite{zamir2022restormer} and UFormer~\cite{wang2022uformer}) on both the Visdrone~\cite{zhu2021detection} and LSRW dataset, which suggests the importance of illumination guidance. Specifically, our enhancement model obtains the highest performance across all three non-referenced metrics (FID~\cite{heusel2017gans}, NIQE, and LOE), which suggests that enhanced images generated by our Mamba-based low-light image enhancement backbone are more perceptual-friendly.
Thirdly, compared with current SOTA Transformer-based methods with illumination estimation,  Retinexformer~\cite{cai2023retinexformer}, our Mamba-based low-light image enhancement backbone obtains $0.66$ dB gains on PSNR and $0.0021$ gains on SSIM on the LSRW dataset. Fourthly, compared with the purely Mamba-based method MambaIR~\cite{guo2024mambair}, our Mamba-based low-light image enhancement backbone archives a $0.54$ performance improvement in the PSNR metric on the LSRW dataset.

\subsubsection{Evaluation on Data Augmentation Strategy}
\begin{table}[ht]
	\centering
	\caption{Investigation on different augmentation strategies. }
 
	\resizebox{0.9\linewidth}{!}{
	\begin{tabular}{c|ccc|cc}
		\toprule[1.pt]
            \multirow{2}{*}{Augmentation Strategy}  & \multicolumn{3}{c|}{\cellcolor{gray!40}\emph{Visdrone Dataset}} & \multicolumn{2}{c}{\cellcolor{gray!40}\emph{LSRW Dataset}} \\ \cline{2-6}	
             
		& \cellcolor{gray!40}FID $\downarrow$ & \cellcolor{gray!40}NIQE $\downarrow$ & \cellcolor{gray!40}LOE $\downarrow$ & \cellcolor{gray!40}PSNR $\uparrow$ & \cellcolor{gray!40}SSIM $\uparrow$ \\ 
		
        \midrule
		Baseline  & 69.85 & 4.153  & 263.5 & 17.96 & 0.7166 \\ 
		  w/ Color Jitter  & 54.97 & 3.894 & 231.2 & 18.98 & 0.7556 \\ 
		  w/ Gaussian Blur  & 48.82 & 3.825 & 226.7 & 19.04 & 0.7678 \\ 
  		w/ Grayscale Conversion & \underline{43.94} & \underline{3.945} & \underline{219.7} & \underline{19.15} & \underline{0.7788} \\ 
    
        \midrule
		All (Ours) & \textbf{36.49}  & \textbf{3.667} & \textbf{204.1} & \textbf{19.73} & \textbf{0.7840} \\
        \bottomrule[1.pt]
	\end{tabular}}

    \label{tab:abl_data_aug}  
\end{table}

We investigate the effectiveness of applying varied strong data augmentation operations in our Semi-LLIE framework and report the evaluation results in Table~\ref{tab:abl_data_aug}. The Baseline refers to our Semi-LLIE, which does not utilize any strong data augmentations. 
Compared to the baseline without any data augmentation, each data augmentation operation (Gaussian blur, grayscale conversion, and color jittering) effectively improves the enhancement performance. Combining all the data augmentation operations, our Semi-LLIE obtains the best performance.


\begin{figure}[htbp]
	\centering
	\includegraphics[width=0.86\linewidth]{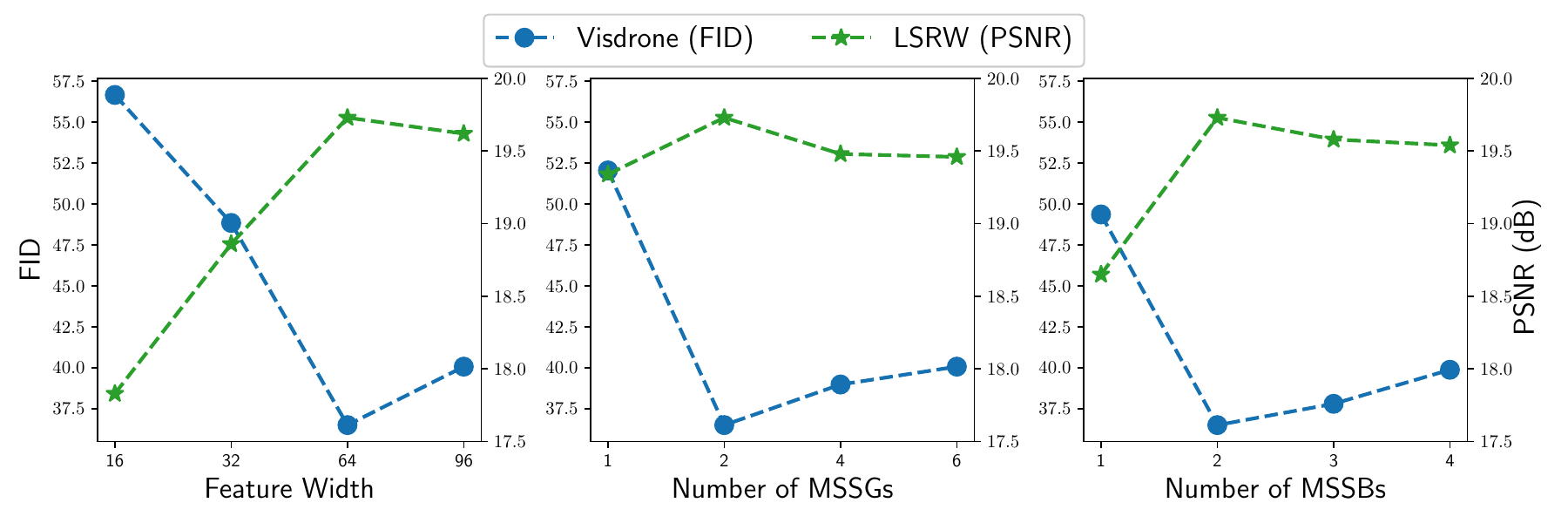}
	\caption[]{Demonstration of essential design choices ablation results for feature width, number of MSSGs of our Mamba-based low-light image enhancement backbone, and number of MSSBs in each MSSG. }
	\label{fig:ablation_design_choices}
        \vspace{-1.0em}
\end{figure}
\subsubsection{Evaluation on Essential Design Choices}
We investigate the effectiveness of model width, the number of MSSGs of our Mamba-based low-light image enhancement backbone, and the number of MSSBs in each MSSG and illustrate the results in Fig.~\ref{fig:ablation_design_choices}. The proposed Semi-LLIE achieves the best FID~\cite{heusel2017gans} value on the Visdrone dataset~\cite{zhu2021detection}, the highest PSNR on the LSRW dataset~\cite{hai2023r2rnet}, and is saturated with the feature width set as $64$. As the number of MSSGs increases, the performance of our Semi-LLIE first improves and then slightly declines, achieving optimal results when the number of MSSGs is set as $2$. For evaluating the number of MSSBs in each MSSG, we find that $2$ leads to the best performance too. 

\begin{table}[ht]
    \centering
    \caption{Object detection results using Faster-RCNN~\cite{ren2015fasterrcnn} on the enhanced images produced by various image enhancement methods.} 

    \resizebox{0.9\columnwidth}{!}{%
    \begin{tabular}{c|c|cccccc}
	\toprule[1.pt]
        \rowcolor{gray!40} Model  & Epochs & AP & AP$_{50}$ & AP$_{75}$ & AP$_{S}$ & AP$_{M}$ & AP$_{L}$  \\
        \midrule
        Original & 12 & 8.0 & 15.7 & 7.2 & 2.9 & 13.2 & 19.5  \\
        RetinexNet~\cite{Chen2018Retinex} & 12 &  6.5 &  13.1 &  5.9 &  2.4 & 11.3 &  19.9  \\
        Restormer~\cite{zamir2022restormer} & 12 &  8.3 &  16.4 &  7.5 &  3.1 & 13.4 &  21.5  \\
        SNR-Net~\cite{xu2022snr} & 12 &  8.2 &  16.1 &  7.3 &  3.1 & 13.8 &  21.0  \\
        Retinexformer~\cite{cai2023retinexformer} & 12 &  8.5 &  17.4 &  7.6 &  3.1 & 13.7 &  21.7  \\
        MambaIR~\cite{guo2024mambair} & 12 &  8.8 &  17.8 &  7.7 &  3.2 & 13.9 &  22.1  \\
        \midrule
        EnGAN~\cite{jiang2021enlightengan} & 12 &  8.2 &  16.0 &  7.4 &  3.1 & 13.7 &  20.9  \\
        SCI~\cite{ma2022sci}   & 12 & 6.8 & 13.6 & 6.0 & 2.5 & 11.6 & 21.3   \\ 
        ZeroDCE~\cite{guo2020zerodce} & 12 & 9.0 & 17.9 & 7.9 & 3.3 & 14.2 & 17.5  \\  
        NeRCo~\cite{yang2023implicit}  & 12 & \underline{9.1} & \underline{18.1} & \underline{8.0} & \underline{3.4} & \underline{14.8} & \underline{24.1}   \\ 
        \bottomrule[1.pt]
        
        Semi-LLIE (Ours)  & 12 & \textbf{9.4} & \textbf{18.3} & \textbf{8.5} & \textbf{3.5} & \textbf{14.9} & \textbf{24.7}  \\
        \bottomrule[1.pt]

    \end{tabular}}
    \centering
    \vspace{-1.5em}
    \label{tab:det_results_on_Visdrone}
    
\end{table}

\section{Pre-Processing for Nighttime Object Detection}

We perform experiments on the nighttime VisDrone dataset~\cite{zhu2021detection} to verify the preprocessing effects of various enhancement algorithms on nighttime object detection. As shown in the last column of Table~\ref{tab:semi-dataset-visdrone}, the nighttime VisDrone dataset~\cite{zhu2021detection} comprises $657$ real testing low-light images, each annotated with bounding boxes for $11$ distinct object categories. In our experiment, the detection models are trained using the $10$ specified categories from the VisDrone dataset~\cite{zhu2021detection}, excluding the "others" category. $526$ images are selected for training and the left $131$ images are for testing. 
We employ Faster-RCNN~\cite{ren2015fasterrcnn} as the detector and it is trained from scratch on the enhanced images produced by distinct pre-trained low-light enhancement methods serving as a preprocessing operation.
The detection results can be viewed as an indirect measure of the semantic preservation ability of different enhancement methods.

The detection results on the VisDrone dataset~\cite{zhu2021detection} are summarized in Table~\ref{tab:det_results_on_Visdrone}, where "Original" refers to the original low-light nighttime images of the VisDrone dataset~\cite{zhu2021detection}.
It can be observed that our Semi-LLIE has achieved a $1.4$ AP improvement compared to the "Original" low-light nighttime images. Enhancing low-light images with our Semi-LLIE as a preprocessing step can substantially boost the effectiveness of the detection model. Besides, nighttime images captured by UAVs present complex illumination conditions, significantly reducing the effectiveness of supervised image enhancement techniques like Retinexfomer~\cite{cai2023retinexformer}. Moreover, previously unsupervised methods (SCI~\cite{ma2022sci} and ZeroDCE~\cite{guo2020zerodce}) achieve sub-optimal performance. The reason is that they focus on enlightening the images by transforming their pixel values according to their illumination characteristics while neglecting to preserve their semantic structure information. In contrast to existing methods, our Semi-LLIE considers pixel-level transformations while harnessing the capabilities of large-scale vision foundation models to preserve semantic information.


\section{Limitations and Future Works}
Although our Semi-LLIE has achieved strong performance, it also shows several limitations. Firstly, our Semi-LLIE requires significant computational resources during training, as the semantic-aware contrastive loss and the RAM-based perceptual loss rely on the large-scale vision-language model RAM~\cite{zhang2024recognize}. 
Secondly, our method exhibits substantial advantages in both performance and computational efficiency over most supervised approaches as shown in Table~\ref{tab:full_referenced_comp_results}. However, compared with some unsupervised methods (SCI~\cite{ma2022sci} and RUAS~\cite{liu2021ruas}), the number of parameters of the proposed Mamba-based low-light enhancement backbone is relatively high. Moreover, the performance potential of the Mamba-based low-light enhancement model is far from being fully realized. 

In the future, we will first investigate more effective yet straightforward consistency regularization techniques for semi-supervised student-teacher learning in low-light image enhancement based on our Semi-LLIE framework.
Then, we will focus on reducing the parameters of the Mamba-based low-light image enhancement backbone and designing more effective Mamba-based image enhancement backbones.
\section{Conclusion}
We propose Semi-LLIE, a semi-supervised framework for low-light image enhancement, to facilitate generating photo-realistic enhancement images. 
Our Semi-LLIE is constructed based on the mean-teacher technique for leveraging paired and unpaired data to boost its generalization ability on low-light images captured at real scenes. The newly designed semantic-aware contrastive loss that acts as a consistency loss can faithfully transfer the illumination distributions learned from the paired data to the unpaired data. 
Besides, it brings implicit semantic information into the learning process of our framework and enables us to produce images with natural colors. To recover realistic textural details, we propose a Mamba-based low-light image enhancement backbone, which can efficiently model global-local pixel relationships and seamlessly collaborate with our Semi-LLIE framework. Moreover, we design a RAM-based perceptual loss to evaluate the distances between positive and negative pairs by leveraging intermediate feature representations extracted from different stages of RAM’s image encoder. Our RAM-based perceptual loss further facilitates the recovery of photo-realistic details originally buried in underexposure regions of low-light images. Experiments on the Visdrone and LSRW datasets have verified the effectiveness and superiority of our Semi-LLIE against other SOTA methods.

\bibliographystyle{IEEEtran}

\bibliography{IEEEabrv, ./reference.bib}

\end{document}